\documentclass{article} 
\usepackage{iclr2027_conference,times}

\usepackage{amsmath,amsfonts,bm}

\def\eqref#1{equation~\ref{#1}}

\def\1{\bm{1}}

\DeclareMathAlphabet{\mathsfit}{\encodingdefault}{\sfdefault}{m}{sl}
\SetMathAlphabet{\mathsfit}{bold}{\encodingdefault}{\sfdefault}{bx}{n}

\usepackage{url}

\usepackage{microtype}
\usepackage{amsmath}
\usepackage{amssymb}
\usepackage{amsfonts}
\usepackage{amsthm}

\usepackage{xcolor}
\usepackage{graphicx}
\usepackage{float}
\usepackage{wrapfig}
\usepackage{subcaption}
\usepackage{colortbl}

\usepackage{booktabs}
\usepackage{array}
\usepackage{tabularx}
\usepackage{threeparttable}
\usepackage{multirow}
\usepackage{multicol}
\usepackage{makecell}
\usepackage{longtable}
\usepackage{ragged2e}
\usepackage{adjustbox}
\usepackage{caption}
\usepackage{calc}
\usepackage{tikz}
\usepackage{algorithm}
\usepackage{hhline}
\definecolor{citeblue}{RGB}{70,110,180}
\usepackage[colorlinks=true, citecolor=citeblue]{hyperref}
\usepackage[most]{tcolorbox}
\usepackage{enumitem}
\usepackage{nicefrac}
\usepackage{listings}
\newcommand{\OA}[1]{\cellcolor{red!10} #1}    
\newcommand{\OB}[1]{\cellcolor{blue!10} #1}   
\newcommand{\EQ}[1]{\cellcolor{orange!10} #1}   

\newcommand{\textOA}[1]{\textcolor{red!70!black}{#1}}     
\newcommand{\textOB}[1]{\textcolor{blue!70!black}{#1}}    
\newcommand{\textEQ}[1]{\textcolor{orange!80!black}{#1}}  

\definecolor{lightblue}{RGB}{225,240,250}
\definecolor{softred}{RGB}{192,0,0}
\definecolor{darkgreen}{RGB}{0,100,70}

\title{Rethinking Faithfulness in LLMs: A Pairwise Context-Sensitive Perspective}

\makeatletter
\renewcommand\@fnsymbol[1]{%
  \ifcase#1 \or \dagger \or \ddagger \or \S \or \P \or **\or \dagger\dagger
  \or \ddagger\ddagger \else \@ctrerr \fi}
\makeatother

\author{
\begin{tabular}{@{}l@{}}
\textbf{Zizhuo Zhang}$^{1}$ \quad
\textbf{Xiong Peng}$^{1}$ \quad
\textbf{Jingwei Sun}$^{1}$ \quad
\textbf{Rong Yao}$^{2}$ \quad
\textbf{Borui Jiang}$^{2}$ \quad
\textbf{Bo Han}$^{1}$\thanks{Correspondence to Bo Han (bhanml@comp.hkbu.edu.hk).}
\end{tabular}
\vspace{0mm} \\
$^{1}$TMLR Group, Department of Computer Science, Hong Kong Baptist University \\ 
$^{2}$Huawei Noah's Ark Lab.
}

\iclrfinalcopy

\begin{document}

\maketitle

\lhead{}

\begin{abstract}
    Large language models (LLMs) are expected to answer questions faithfully based on the provided context, abstaining when the context information is insufficient to answer the questions. Existing faithfulness evaluations typically assess each question-context instance in isolation; however, such instance-level evaluation fails to capture a fundamental requirement of faithful behavior: the ability to adapt model responses to changes in available contexts. In particular, a model should provide correct answers when sufficient evidence is present and abstain when it is not. In this work, we propose a Pairwise Faithfulness Benchmark (PFaithBench) that evaluates whether a model can switch between answering and abstaining for the same question under supporting versus non-supporting contexts. Our evaluations across thirty-nine models with seven model families demonstrate that faithfulness fundamentally involves a trade-off between answering and abstaining, and that most current models exhibit a strong bias toward answering, with most faithfulness errors arising from over-answering, i.e., models tend to fabricate a response even when the provided context is insufficient. We further conduct a series of studies on faithfulness training under different data constructions. Our results show that training outcomes are highly sensitive to the specific composition of answering and abstaining data. Constructing answering and abstaining data from mismatched sources can cause models to rely on dataset-specific shortcuts rather than actual context sufficiency. Moreover, increasing answer-supervised data improves answering performance but exacerbates over-answering, while increasing abstaining data reduces hallucination but leads to over-abstention. The code and data are released at~\url{https://github.com/tmlr-group/PFaithBench}.
\end{abstract}

\begin{figure}[h]
  \centering
  \vspace{-2mm}
  \includegraphics[width=0.99\textwidth]{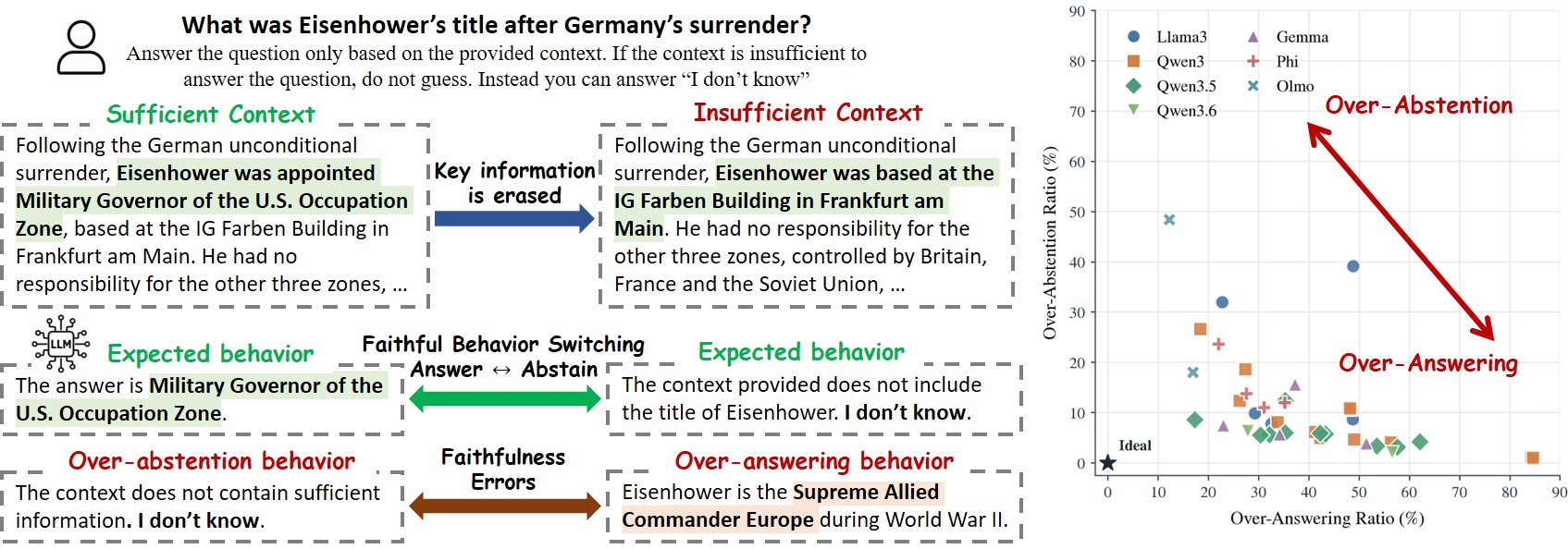}
  \vspace{-1mm}
  \caption{\textbf{Overview of pairwise faithfulness behavior}. \textit{Left:} Given the same question with sufficient and insufficient contexts, a faithful model should exhibit context-sensitive behavior: answering when sufficient evidence is available and abstaining when key information is missing. Over-abstention and over-answering are two sources of faithfulness errors. \textit{Right:} Evaluations across seven families with thirty-nine models: most models exhibit bias toward over-answering, where one point is one model.}
  \vspace{-2mm}
  \label{Fig:Demo}
\end{figure}

\section{Introduction}
\label{Sec:Introduction}

Large language models (LLMs) are increasingly deployed in knowledge-intensive applications~\citep{bhathena2026mm,kalra2024hypa}, often with external contexts provided through retrieval-augmented generation (RAG)~\citep{gao2023RAGSurvey} or domain-specific knowledge bases~\citep{edge2024graphrag}. In such setting, a fundamental requirement for reliable deployment of LLMs is \emph{faithfulness}: models should generate responses that are strictly grounded in the provided context, answering when sufficient evidence context is available and abstaining when it is not~\citep{huang2025LLMhallucinationsurvey, mingfaitheval, bi2025contextDPO}. However, LLM may fabricate fluent but unsupported responses when context is insufficient, leading to over-answering and hallucinations, or abstain despite sufficient context, resulting in over-abstention~\citep{joren2025sufficient}. This raises a fundamental question: \textit{how should LLM behavior be characterized when the provided context is sufficient versus insufficient?}

\par
\begin{figure}[t]
  \centering
  \includegraphics[width=0.99\textwidth]{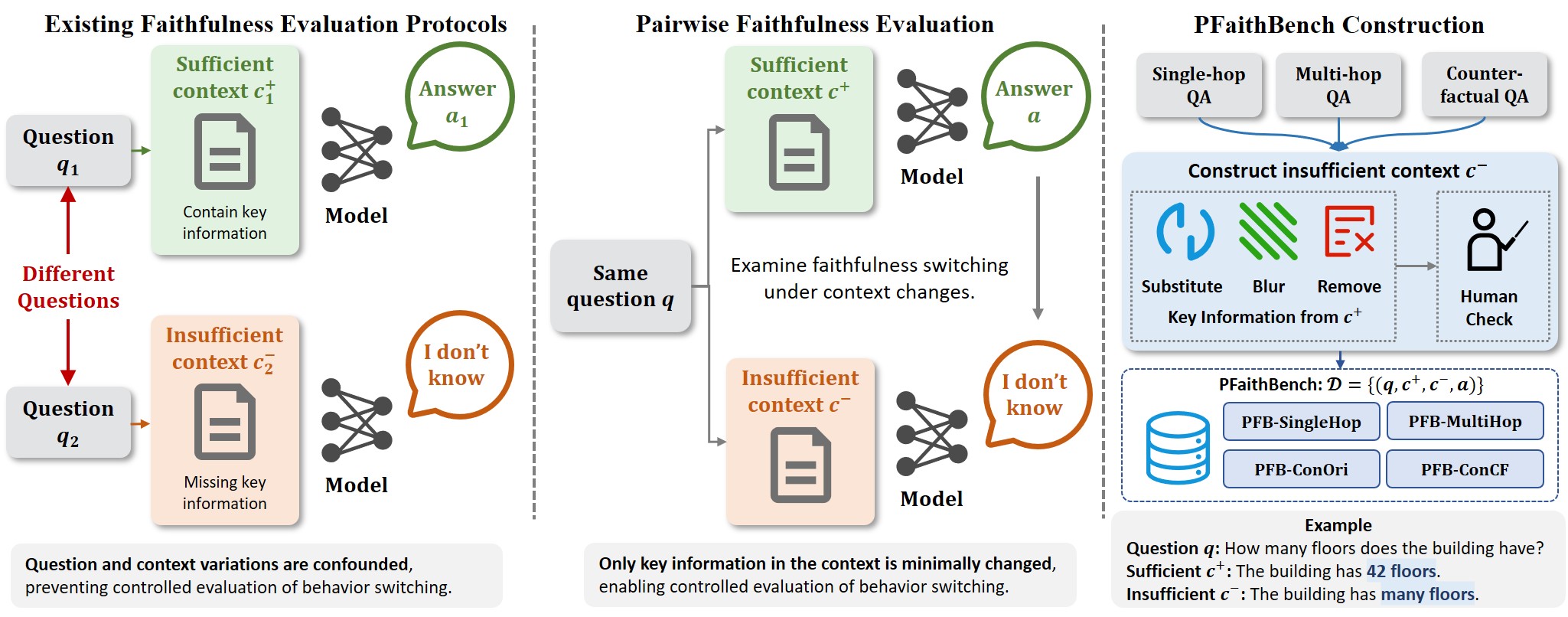}
  \vspace{-1mm}
  \caption{\textbf{Pairwise faithfulness evaluation and PFaithBench construction.}  
  \textit{Left:} Existing evaluation protocols assess individual question-context instances, confounding question and context variations. \textit{Middle:} Pairwise evaluation keeps the question fixed while varying contextual sufficiency, enabling controlled assessment of faithfulness switching. \textit{Right:} PFaithBench is constructed from multi-hop, single-hop, and counterfactual QA by modifying key information in sufficient contexts.
  }
  \vspace{-5mm}
  \label{Fig:overview}
\end{figure}

Existing evaluation protocols typically assess faithfulness at the level of individual question-context instances. Prior work considers a range of contextual conditions, from sufficient context for evaluating answer correctness~\citep{adlakha2024evaluating,ho20202wiki} to insufficient context for assessing abstaining behavior~\citep{mingfaitheval}, as well as settings involving noisy~\citep{chen2024benchmarking}, counterfactual~\citep{bi2025contextDPO}, or time-related~\citep{chen2021TimeQA} context. However, as illustrated in Fig.~\ref{Fig:overview}, these conditions are usually evaluated independently and on different sets of questions. This entangles question variation with context variation, preventing a controlled assessment of whether models appropriately switch between answering and abstaining as contextual sufficiency changes.

\par
In this work, we propose a \textit{Pairwise Faithfulness Benchmark} (PFaithBench), which characterizes model behavior under sufficient and insufficient contextual conditions for the same question. As illustrated in Fig.~\ref{Fig:overview}, by keeping the question fixed while varying contextual sufficiency, PFaithBench enables a controlled assessment of whether models appropriately switch between answering and abstaining, without confounding from the variation of questions. We characterize two primary types of faithfulness errors: (i) \emph{Over-answering} (OverAns), where models answer despite insufficient context, and (ii) \emph{Over-abstention} (OverAbs), where models abstain despite sufficient context. As shown in Fig.~\ref{Fig:Demo}, our evaluations across 39 models from seven families find that most existing models exhibit a bias toward over-answering, and a portion of over-answering stems from internal knowledge leakage, where models rely on internal parametric knowledge beyond the provided context.

\par
Additionally, we conduct a series of data-centric analyses to investigate how training data composition affects faithfulness behaviors of models. We find that constructing answering and abstaining data from mismatched sources can induce dataset-specific shortcuts, causing models to rely on source-specific signals rather than contextual sufficiency. We further observe a clear trade-off between answering and abstaining supervision: increasing answering data improves answering performance but exacerbates over-answering, whereas increasing abstention data reduces unsupported answering but leads to over-abstention. These findings highlight the importance of proper context coupling and a balanced composition of answering and abstaining data for effective faithfulness training.

\par
Our main contributions of this work are summarized as follows:
\begin{itemize}[leftmargin=1.3em, labelsep=0.5em, nosep]
\item We propose PFaithBench, which assesses whether models can appropriately switch between answering and abstaining for the same question under sufficient and insufficient contexts.
\item  We identify two primary sources of faithfulness errors: over-answering and over-abstention, demonstrating a prevalent tendency toward over-answering among existing models.
\item We conduct data-centric analyses, highlighting the importance of proper context coupling and balanced answering-abstention data composition for faithfulness training.
\end{itemize}

\begin{table}[t]
\centering
\caption{\textbf{Evaluation of Over-answering and Over-abstention on PFaithBench across diverse LLMs.} \textOA{Lightred} indicates OverAns exceeds OverAbs by more than 5\%, \textOB{lightblue} indicates OverAbs exceeds OverAns by more than 5\%, and \textEQ{lightorange} indicates an absolute difference within 5\%.}
\label{Tble:OverAns_OverAbs}
\vspace{-3mm}
\resizebox{\textwidth}{!}{
\begin{threeparttable}
\begin{tabular}{ll|cc|cc|cc|cc|cc}
    \toprule[1.5pt]
    \multirow{2}*{\textbf{Family}} & \multirow{2}*{\textbf{LLM}} & \multicolumn{2}{c}{\textbf{PFB-MultiHop}} & \multicolumn{2}{c}{\textbf{PFB-SingleHop}} & \multicolumn{2}{c}{\textbf{PFB-ConOri}} & \multicolumn{2}{c}{\textbf{PFB-ConCF}} & \multicolumn{2}{c}{\textbf{Avg.}} \\
    \cmidrule(lr){3-4} \cmidrule(lr){5-6} \cmidrule(lr){7-8} \cmidrule(lr){9-10} \cmidrule(lr){11-12}
    ~ & ~ & \textbf{OverAns} $\downarrow$ & \textbf{OverAbs} $\downarrow$ & \textbf{OverAns} $\downarrow$ & \textbf{OverAbs} $\downarrow$ & \textbf{OverAns} $\downarrow$ & \textbf{OverAbs} $\downarrow$ & \textbf{OverAns} $\downarrow$ & \textbf{OverAbs} $\downarrow$ & \textbf{OverAns} $\downarrow$ & \textbf{OverAbs} $\downarrow$ \\
    \midrule
    \multirow{5}*{\textbf{Llama3}} & Llama-3.2-1B-Instruct & \OA{53.90} & 35.50 & \OA{52.76} & 40.81 & \OA{44.25} & 37.04 & \EQ{44.37} & \EQ{43.11} & \OA{48.82} & 39.12 \\
    ~ & Llama-3.2-3B-Instruct & \OA{37.40} & 23.90 & 12.99 & \OB{43.18} & \EQ{19.47} & \EQ{23.51} & 21.11 & \OB{37.42} & 22.74 & \OB{32.00} \\
    ~ & Llama-3.1-8B-Instruct & \OA{56.60} & 6.50 & \OA{45.41} & 6.04 & \OA{45.89} & 3.79 & \OA{47.03} & 18.46 & \OA{48.73} & 8.70 \\
    ~ & Llama-3.1-70B-Instruct & \OA{37.20} & 7.40 & \OA{21.00} & 5.64 & \OA{28.19} & 7.71 & \OA{30.47} & 18.58 & \OA{29.21} & 9.83 \\
    ~ & Llama-3.3-70B-Instruct & \OA{37.20} & 7.20 & \OA{27.69} & 3.54 & \OA{32.24} & 6.45 & \OA{33.12} & 13.78 & \OA{32.56} & 7.74 \\
    \midrule
    \multirow{12}*{\textbf{Qwen3}} & Qwen3-0.6B (NoThink) & \OA{57.00} & 17.80 & \OA{49.87} & 11.29 & \OA{44.50} & 5.44 & \OA{41.85} & 7.96 & \OA{48.30} & 10.62 \\
    ~ & Qwen3-1.7B (NoThink) & \OA{57.00} & 17.90 & \OA{49.48} & 10.76 & \OA{43.11} & 5.69 & \OA{43.36} & 8.98 & \OA{48.24} & 10.83 \\
    ~ & Qwen3-4B (NoThink) & \OA{32.90} & 25.80 & \EQ{20.34} & \EQ{18.24} & \OA{28.19} & 8.22 & \OA{27.94} & 22.12 & \OA{27.34} & 18.60 \\
    ~ & Qwen3-8B (NoThink) & 23.20 & \OB{31.70} & \OA{20.87} & 15.09 & \EQ{14.79} & \EQ{13.91} & 14.92 & \OB{45.76} & 18.44 & \OB{26.62} \\
    ~ & Qwen3-14B (NoThink) & \OA{36.70} & 13.50 & \OA{22.31} & 6.69 & \OA{21.11} & 6.32 & \EQ{24.78} & \EQ{22.88} & \OA{26.23} & 12.35 \\
    ~ & Qwen3-32B (NoThink) & \OA{41.20} & 7.10 & \OA{35.04} & 2.76 & \OA{27.69} & 5.31 & \OA{31.35} & 17.19 & \OA{33.82} & 8.09 \\
    \cmidrule(lr){2-12}
    ~ & Qwen3-0.6B (Think) & \OA{84.90} & 2.10 & \OA{88.06} & 1.05 & \OA{80.78} & 1.14 & \OA{83.44} & 0.51 & \OA{84.30} & 1.20 \\
    ~ & Qwen3-1.7B (Think) & \OA{85.60} & 1.90 & \OA{89.11} & 0.52 & \OA{81.80} & 0.63 & \OA{82.17} & 1.01 & \OA{84.67} & 1.02 \\
    ~ & Qwen3-4B (Think) & \OA{46.30} & 3.70 & \OA{41.34} & 2.76 & \OA{40.46} & 4.55 & \OA{41.21} & 8.85 & \OA{42.33} & 4.96 \\
    ~ & Qwen3-8B (Think) & \OA{46.70} & 5.20 & \OA{35.30} & 3.15 & \OA{40.71} & 4.93 & \OA{42.23} & 11.38 & \OA{41.23} & 6.16 \\
    ~ & Qwen3-14B (Think) & \OA{45.80} & 3.70 & \OA{44.49} & 1.71 & \OA{49.05} & 3.41 & \OA{56.89} & 9.73 & \OA{49.06} & 4.64 \\
    ~ & Qwen3-32B (Think) & \OA{52.00} & 3.30 & \OA{47.90} & 1.31 & \OA{60.30} & 3.79 & \OA{65.36} & 7.71 & \OA{56.39} & 4.03 \\
    \midrule
    \multirow{10}*{\textbf{Qwen3.5}} & Qwen3.5-0.8B (NoThink) & \OA{53.80} & 12.00 & \OA{36.22} & 15.22 & \OA{26.17} & 9.61 & \OA{24.91} & 12.77 & \OA{35.27} & 12.40 \\
    ~ & Qwen3.5-2B (NoThink) & \OA{60.50} & 5.30 & \OA{46.46} & 4.33 & \OA{31.23} & 5.18 & \OA{35.02} & 8.22 & \OA{43.30} & 5.76 \\
    ~ & Qwen3.5-4B (NoThink) & \OA{52.00} & 4.60 & \OA{31.36} & 2.62 & \OA{29.46} & 5.06 & \OA{28.45} & 11.76 & \OA{35.32} & 6.01 \\
    ~ & Qwen3.5-9B (NoThink) & \OA{47.00} & 3.80 & \OA{30.84} & 2.23 & \OA{26.04} & 5.94 & \OA{24.65} & 10.37 & \OA{32.13} & 5.58 \\
    ~ & Qwen3.5-27B (NoThink) & \OA{48.80} & 3.20 & \OA{27.82} & 1.57 & \OA{22.00} & 5.69 & \OA{22.88} & 11.76 & \OA{30.38} & 5.56 \\
    \cmidrule(lr){2-12}
    ~ & Qwen3.5-0.8B (Think) & \OA{76.80} & 2.60 & \OA{62.47} & 3.02 & \OA{55.37} & 4.42 & \OA{53.86} & 6.83 & \OA{62.12} & 4.22 \\
    ~ & Qwen3.5-2B (Think) & \OA{63.80} & 3.00 & \OA{37.80} & 2.89 & \OA{33.50} & 6.57 & \OA{34.01} & 11.25 & \OA{42.28} & 5.93 \\
    ~ & Qwen3.5-4B (Think) & \OA{74.50} & 1.80 & \OA{55.25} & 0.79 & \OA{49.94} & 3.92 & \OA{50.82} & 6.19 & \OA{57.63} & 3.18 \\
    ~ & Qwen3.5-9B (Think) & \OA{32.20} & 5.50 & \OA{18.11} & 5.12 & \EQ{8.47} & \EQ{8.09} & 10.49 & \OB{15.80} & \OA{17.32} & 8.63 \\
    ~ & Qwen3.5-27B (Think) & \OA{80.10} & 1.10 & \OA{50.52} & 0.92 & \OA{40.08} & 4.42 & \OA{43.36} & 6.95 & \OA{53.52} & 3.35 \\
    \midrule
    \multirow{2}*{\textbf{Qwen3.6}} & Qwen3.6-27B (NoThink) & \OA{41.30} & 4.40 & \OA{26.25} & 1.71 & \OA{22.38} & 6.57 & \OA{21.49} & 12.90 & \OA{27.85} & 6.39 \\
    ~ & Qwen3.6-27B (Think) & \OA{79.10} & 0.90 & \OA{54.46} & 0.26 & \OA{45.26} & 3.29 & \OA{47.41} & 4.05 & \OA{56.56} & 2.12 \\
    \midrule
    \multirow{4}*{\textbf{Gemma}} & Gemma-2-2B-It & \OA{44.00} & 30.90 & \OA{29.53} & 13.12 & \OA{38.56} & 6.70 & \OA{36.92} & 11.76 & \OA{37.25} & 15.62 \\
    ~ & Gemma-2-9B-It & \OA{70.90} & 5.00 & \OA{42.52} & 2.76 & \OA{46.02} & 2.91 & \OA{46.27} & 4.80 & \OA{51.43} & 3.87 \\
    ~ & Gemma-2-27B-It & \OA{53.40} & 6.40 & \OA{30.31} & 2.49 & \OA{23.77} & 4.93 & \OA{29.08} & 8.85 & \OA{34.14} & 5.67 \\
    ~ & Gemma-4-31B-it & \OA{38.80} & 5.60 & \OA{19.95} & 2.23 & \OA{15.68} & 8.85 & \EQ{17.57} & \EQ{13.27} & \OA{23.00} & 7.49 \\
    \midrule
    \multirow{4}*{\textbf{Phi}} & Phi-3-mini-128k-instruct & \OA{34.80} & 26.40 & 17.85 & \OB{29.13} & \EQ{17.57} & \EQ{13.02} & 17.95 & \OB{25.92} & \EQ{22.04} & \EQ{23.62} \\
    ~ & Phi-3.5-mini-instruct & \OA{55.30} & 14.90 & \OA{25.85} & 16.14 & \OA{30.09} & 4.93 & \OA{29.58} & 12.01 & \OA{35.21} & 12.00 \\
    ~ & Phi-4-mini-instruct & \OA{38.70} & 16.10 & \OA{22.57} & 16.40 & \OA{23.26} & 7.33 & \OA{25.92} & 15.17 & \OA{27.61} & 13.75 \\
    ~ & Phi-4 & \OA{42.30} & 6.30 & \OA{30.18} & 4.99 & \OA{24.78} & 8.47 & \EQ{27.18} & \EQ{24.27} & \OA{31.11} & 11.01 \\
    \midrule
    \multirow{2}*{\textbf{Olmo}} & Olmo-3-7B-Instruct & 19.60 & \OB{52.60} & 6.69 & \OB{54.07} & 10.37 & \OB{30.85} & 12.26 & \OB{56.01} & 12.23 & \OB{48.38} \\
    ~ & Olmo-3.1-32B-Instruct & \EQ{21.90} & \EQ{20.20} & \EQ{13.25} & \EQ{14.44} & \EQ{14.03} & \EQ{10.49} & 18.46 & \OB{27.05} & \EQ{16.91} & \EQ{18.05} \\
    \bottomrule[1.5pt]
\end{tabular}
\end{threeparttable}}
\vspace{-5mm}
\end{table}

\section{Related Work}
\label{Sec:RelatedWork}
\textbf{Faithfulness evaluation.} 
Faithfulness has emerged as a key requirement for LLMs deployed in information-sensitive RAG systems~\citep{lewis2020retrieval,gao2023RAGSurvey}, where models are expected to strictly ground responses in the provided context rather than parametric knowledge~\citep{ji2023survey}. Representative benchmarks include FaithEval~\citep{mingfaitheval}, which examines unanswerable, inconsistent, and counterfactual contexts; RGB~\citep{chen2024benchmarking}, which investigates robustness to noisy retrieval; ClashEval~\citep{wu2024clasheval} and ContextDPO~\citep{bi2025contextDPO}, which study conflicts between internal knowledge and external evidence; TimeQA~\citep{chen2021TimeQA}, which constructs time-sensitive contexts; and 2WikiMultiHopQA~\citep{ho20202wiki} and HotpotQA~\citep{yang2018hotpotqa}, which focus on answer correctness in multi-hop QA. However, existing evaluations typically assess different questions under different contextual conditions, confounding question and context variations and obscuring models' ability to switch between answering and abstaining. This motivates our pairwise evaluation, which constructs paired sufficient and insufficient contexts for the same question.

\textbf{Faithfulness training.}
Existing studies have explored various approaches to improve faithfulness, including context-grounded fine-tuning, such as RA-DIT~\citep{lin2023ra} and RAFT~\citep{zhang2024raft}; abstention-aware supervision, such as R-Tuning~\citep{zhang2024RTuning}, Alignment for Honesty~\citep{yang2024alignment}, and CanAISayNo~\citep{cheng2024can}; preference optimization, such as ContextDPO~\citep{bi2025contextDPO}; and reinforcement learning, such as FaithRL~\citep{gui2026FaithRL}, KnowRL~\citep{ren2025knowrl}, and Faithfulness-Aware Step-Level RL~\citep{nie2026stop}. Despite these advances, they primarily focus on teaching models to exhibit appropriate behaviors under specific contextual conditions. In this work, we adopt the widely used GRPO~\citep{shao2024deepseekmath} as our training algorithm and conduct a series of data-centric analyses to investigate how training data construction and composition affect context-sensitive answering and abstaining behaviors of models.

\section{Pairwise Faithfulness Evaluation}
To begin, we formalize the pairwise faithfulness evaluation (Sec.~\ref{Sec:Problem_Def}), introduce the corresponding evaluation metrics (Sec.~\ref{Sec:Metrics_PairFaith}), and describe the construction of PFaithBench (Sec.~\ref{Sec:PFaithBench_Contruct}). 

\subsection{Problem Definition}
\label{Sec:Problem_Def}
Pairwise faithfulness evaluation considers question answering (QA) under varying contextual conditions. Given a question $q$ and its ground-truth answer $a$, let $c^+$ and $c^-$ denote two types of contexts associated with $q$: a sufficient context $c^+$ that contains supported key information to answer the question, and an insufficient context $c^-$ lacks such information. Given a model $\pi_\theta$, which takes a question-context pair $(q,c)$ as input and samples a response $y\sim \pi_\theta(\cdot|q,c)$, pairwise faithfulness requires the model to exhibit context-sensitive behaviors as follows:
\begin{equation}
y^+ \sim \pi_\theta(\cdot|q, c^+) \rightarrow \text{answer correctly}, \qquad
y^- \sim \pi_\theta(\cdot|q, c^-) \rightarrow \text{abstain appropriately}.
\end{equation}
That is, the model should generate an answer when sufficient evidence is available, and abstain (e.g., by responding ``I don't know") when the context is insufficient. Ideally, a perfectly faithful model should appropriately switch between these two behaviors when $c^+$ and $c^-$ differ only in the key information required to answer the question. We formalize this setting as \emph{pairwise faithfulness evaluation}, where each question $q$ is associated with a pair of contexts $(c^+, c^-)$, with $c^-$ constructed by minimally modifying the key information in $c^+$. Accordingly, each data is represented as a tuple $(q, c^+, c^-, a)$, where $a$ denotes the ground-truth answer to $q$. The ground-truth answer $a$ enables us to assess response correctness under sufficient context and identify correct answers generated despite insufficient context, potentially indicating reliance on internal parametric knowledge.

\par
Under this formulation, model responses can be categorized into three types: (i) abstention ($\mathrm{IDK}$), where the model chooses not to answer the question (e.g., by responding ``I don't know''); (ii) correct ($\mathrm{C}$), where the response matches the ground-truth $a$; and (iii) wrong ($\mathrm{W}$), where the response does not match the ground-truth $a$ and lacks sufficient supporting evidence from the provided context.

\begin{figure}[t]
  \centering
  \includegraphics[width=0.99\textwidth]{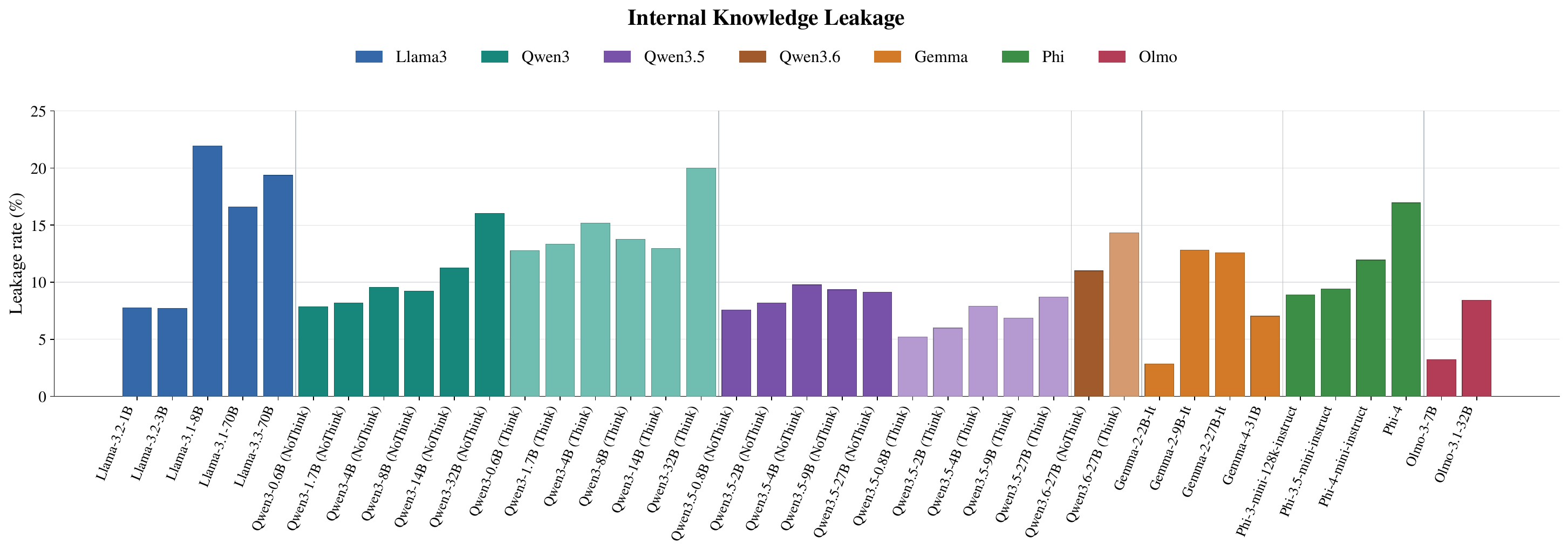}
  \vspace{-2mm}
  \caption{\textbf{Internal knowledge leakage across models.} Leakage measures the extent to which models produce answers by relying on internal parametric knowledge rather than the provided context. Colors indicate different model families, and model sizes increase from left to right within each family. }
  \vspace{-5mm}
  \label{Fig:Leakage}
\end{figure}

\subsection{Metrics of Pairwise Faithfulness}
\label{Sec:Metrics_PairFaith}
Based on the response categories $y\in \{\mathrm{IDK}, \mathrm{C}, \mathrm{W}\}$ of the model, beyond simple accuracy, a set of metrics can be defined to quantify pairwise faithfulness behavior as follows:

\par
\textbf{Over-answering vs. over-abstention.} \emph{Over-answering} (OverAns) measures the tendency of the model to produce a non-abstaining response under insufficient context, while \emph{Over-abstention} (OverAbs) measures the tendency of the model to abstain despite sufficient context:
\begin{equation}
\mathrm{OverAns} = \mathbb{P}(y^- \in \{\mathrm{C}, \mathrm{W}\}), \qquad \mathrm{OverAbs} = \mathbb{P}(y^+ = \mathrm{IDK}).
\end{equation}
These two metrics capture the primary sources of faithfulness errors, reflecting the opposing tendencies of the model to answer too aggressively or abstain too conservatively.

\par
\textbf{Internal knowledge leakage.} 
We quantify the extent to which models rely on internal parametric knowledge rather than the provided context by measuring the probability that a model produces the ground-truth answer despite insufficient context:
\begin{equation}
\mathrm{Leakage} = \mathbb{P}(y^- = \mathrm{C}).
\end{equation}
Ideally, a faithful model should follow the instruction of answering the question strictly based on the provided context rather than its internal parametric knowledge. However, models may still rely on memorized knowledge to answer questions when contextual evidence is insufficient~\citep{mingfaitheval,gui2026FaithRL}, a phenomenon we refer to as \emph{internal knowledge leakage}.

\par
\textbf{Pairwise faithfulness accuracy.} 
We define two accuracy metrics to capture pairwise faithfulness at different levels of strictness: \emph{Decision Faithfulness Accuracy} (DFAcc) and \emph{Strict Faithfulness Accuracy} (SFAcc). Specifically, DFAcc measures whether the model appropriately switches its behaviors between answering and abstaining under sufficient and insufficient contexts, while SFAcc additionally requires the answer to be correct under sufficient context:
\begin{equation}
    \mathrm{DFAcc} = \mathbb{P}(y^+ \in \{\mathrm{C}, \mathrm{W}\}, \; y^- = \mathrm{IDK}), \qquad \mathrm{SFAcc} = \mathbb{P}(y^+ = \mathrm{C}, \; y^- = \mathrm{IDK}).
\end{equation}
Compared to DFAcc, SFAcc imposes a stricter requirement by considering answer correctness. The gap between DFAcc and SFAcc reflects the model's inability to produce correct answers under sufficient context despite making appropriate answering and abstaining decisions.

\begin{table}[t]
\centering
\caption{\textbf{Evaluation of pairwise accuracy DFAcc and SFAcc on PFaithBench across diverse LLMs.} The $\mathrm{Gap}=\mathrm{DFAcc}-\mathrm{SFAcc}$ indicates the discrepancy between appropriate answer-or-abstain decisions and correct answering. \textOA{Lightred} indicates large gaps (Gap $\geq 10$), while \textOB{lightblue} indicates small gaps (Gap $< 10$). The best results in each family are highlighted in \textbf{bold}.}
\label{Tble:DFAcc_SFAcc}
\vspace{-3mm}
\resizebox{\textwidth}{!}{
\begin{threeparttable}
\begin{tabular}{ll|cc|cc|cc|cc|ccc}
\toprule[1.5pt]
\multirow{2}*{\textbf{Family}} & \multirow{2}*{\textbf{LLM}} & \multicolumn{2}{c}{\textbf{PFB-MultiHop}} & \multicolumn{2}{c}{\textbf{PFB-SingleHop}} & \multicolumn{2}{c}{\textbf{PFB-ConOri}} & \multicolumn{2}{c}{\textbf{PFB-ConCF}} & \multicolumn{3}{c}{\textbf{Avg.}} \\
\cmidrule(lr){3-4} \cmidrule(lr){5-6} \cmidrule(lr){7-8} \cmidrule(lr){9-10} \cmidrule(lr){11-13}
~ & ~ & \textbf{DFAcc} $\uparrow$ & \textbf{SFAcc} $\uparrow$ & \textbf{DFAcc} $\uparrow$ & \textbf{SFAcc} $\uparrow$ & \textbf{DFAcc} $\uparrow$ & \textbf{SFAcc} $\uparrow$ & \textbf{DFAcc} $\uparrow$ & \textbf{SFAcc} $\uparrow$ & \textbf{DFAcc} $\uparrow$ & \textbf{SFAcc} $\uparrow$ & \textbf{Gap} $\downarrow$ \\
\midrule
\multirow{5}*{\textbf{Llama3}} & Llama-3.2-1B-Instruct & 21.50 & 5.60 & 17.06 & 6.69 & 27.81 & 7.21 & 26.68 & 3.79 & 23.26 & 5.82 & \OA{17.44} \\
~ & Llama-3.2-3B-Instruct & 42.20 & 26.00 & 46.19 & 25.85 & 59.29 & 43.87 & 47.91 & 15.93 & 48.90 & 27.91 & \OA{20.99} \\
~ & Llama-3.1-8B-Instruct & 37.70 & 30.00 & 49.87 & 39.50 & 50.95 & 45.13 & 42.10 & 19.09 & 45.15 & 33.43 & \OA{11.72} \\
~ & Llama-3.1-70B-Instruct & \textbf{56.10} & \textbf{50.00} & \textbf{73.62} & \textbf{68.90} & \textbf{64.85} & \textbf{59.04} & 54.61 & 22.50 & \textbf{62.30} & \textbf{50.11} & \OA{12.19} \\
~ & Llama-3.3-70B-Instruct & 56.00 & 49.30 & 69.16 & 64.70 & 62.20 & 57.77 & \textbf{55.25} & \textbf{26.30} & 60.65 & 49.52 & \OA{11.13} \\
\midrule
\multirow{12}*{\textbf{Qwen3}} & Qwen3-0.6B (NoThink) & 29.80 & 12.30 & 40.55 & 28.35 & 51.20 & 42.86 & 51.45 & 25.54 & 43.25 & 27.26 & \OA{15.99} \\
~ & Qwen3-1.7B (NoThink) & 28.40 & 11.40 & 41.47 & 26.90 & 52.47 & 41.97 & 49.30 & 24.78 & 42.91 & 26.26 & \OA{16.65} \\
~ & Qwen3-4B (NoThink) & 44.10 & 32.90 & 63.12 & 56.56 & 63.72 & 59.29 & 55.12 & 29.58 & 56.52 & 44.58 & \OA{11.93} \\
~ & Qwen3-8B (NoThink) & 47.50 & 37.90 & 64.83 & 60.63 & 71.81 & 69.15 & 46.14 & 34.64 & 57.57 & 50.58 & \OB{6.99} \\
~ & Qwen3-14B (NoThink) & 50.30 & 41.90 & \textbf{71.13} & \textbf{69.03} & \textbf{72.69} & \textbf{69.79} & \textbf{57.14} & \textbf{41.21} & \textbf{62.82} & \textbf{55.48} & \OB{7.33} \\
~ & Qwen3-32B (NoThink) & \textbf{52.80} & \textbf{46.80} & 62.60 & 60.50 & 67.26 & 63.84 & 55.50 & 35.65 & 59.54 & 51.70 & \OB{7.84} \\
\cmidrule(lr){2-13}
~ & Qwen3-0.6B (Think) & 14.50 & 8.40 & 11.42 & 9.58 & 18.58 & 16.43 & 16.43 & 9.86 & 15.23 & 11.07 & \OB{4.17} \\
~ & Qwen3-1.7B (Think) & 13.30 & 7.40 & 10.89 & 9.19 & 17.95 & 15.17 & 17.70 & 10.87 & 14.96 & 10.66 & \OB{4.30} \\
~ & Qwen3-4B (Think) & 50.90 & 42.00 & 56.43 & 52.36 & \textbf{55.50} & 50.32 & \textbf{51.71} & \textbf{32.49} & 53.63 & 44.29 & \OB{9.34} \\
~ & Qwen3-8B (Think) & 49.20 & 41.80 & \textbf{62.07} & \textbf{57.87} & 55.12 & \textbf{50.95} & 50.70 & 31.23 & \textbf{54.27} & \textbf{45.46} & \OB{8.81} \\
~ & Qwen3-14B (Think) & \textbf{51.30} & \textbf{42.80} & 53.94 & 50.39 & 48.04 & 42.86 & 38.56 & 24.02 & 47.96 & 40.02 & \OB{7.94} \\
~ & Qwen3-32B (Think) & 45.60 & 39.50 & 51.05 & 48.82 & 36.54 & 33.88 & 30.34 & 18.08 & 40.88 & 35.07 & \OB{5.81} \\
\midrule
\multirow{10}*{\textbf{Qwen3.5}} & Qwen3.5-0.8B (NoThink) & 35.80 & 22.40 & 50.39 & 30.45 & 65.49 & 55.63 & 64.10 & 30.21 & 53.94 & 34.67 & \OA{19.27} \\
~ & Qwen3.5-2B (NoThink) & 35.40 & 26.10 & 50.00 & 39.63 & 64.48 & 56.01 & 58.91 & 31.35 & 52.20 & 38.27 & \OA{13.92} \\
~ & Qwen3.5-4B (NoThink) & 43.90 & 36.30 & 66.27 & 60.37 & 65.99 & 62.45 & 62.07 & 33.38 & 59.56 & 48.12 & \OA{11.44} \\
~ & Qwen3.5-9B (NoThink) & \textbf{49.80} & \textbf{42.60} & 67.19 & 64.57 & 68.27 & 63.34 & 66.37 & \textbf{39.57} & 62.91 & 52.52 & \OA{10.39} \\
~ & Qwen3.5-27B (NoThink) & 48.30 & 40.80 & \textbf{70.60} & \textbf{68.64} & \textbf{72.44} & \textbf{68.65} & \textbf{66.75} & 37.04 & \textbf{64.52} & \textbf{53.78} & \OA{10.74} \\
\cmidrule(lr){2-13}
~ & Qwen3.5-0.8B (Think) & 21.50 & 13.00 & 35.17 & 27.82 & 41.21 & 34.13 & 42.10 & 21.74 & 35.00 & 24.18 & \OA{10.82} \\
~ & Qwen3.5-2B (Think) & 33.90 & 27.80 & 59.58 & 54.46 & 60.94 & 56.26 & 56.51 & 31.86 & 52.73 & 42.59 & \OA{10.14} \\
~ & Qwen3.5-4B (Think) & 24.10 & 18.30 & 44.23 & 40.81 & 46.65 & 42.98 & 44.88 & 26.30 & 39.96 & 32.10 & \OB{7.87} \\
~ & Qwen3.5-9B (Think) & \textbf{63.10} & \textbf{55.30} & \textbf{76.77} & \textbf{74.15} & \textbf{83.57} & \textbf{79.90} & \textbf{75.47} & \textbf{48.17} & \textbf{74.73} & \textbf{64.38} & \OA{10.35} \\
~ & Qwen3.5-27B (Think) & 19.10 & 13.00 & 48.69 & 45.93 & 56.13 & 51.20 & 51.45 & 28.45 & 43.84 & 34.64 & \OB{9.20} \\
\midrule
\multirow{2}*{\textbf{Qwen3.6}} & Qwen3.6-27B (NoThink) & \textbf{54.40} & \textbf{47.00} & \textbf{72.05} & \textbf{69.42} & \textbf{71.43} & \textbf{67.38} & \textbf{67.26} & \textbf{36.92} & \textbf{66.28} & \textbf{55.18} & \OA{11.10} \\
~ & Qwen3.6-27B (Think) & 20.60 & 16.60 & 45.41 & 44.09 & 51.83 & 48.42 & 49.05 & 25.92 & 41.72 & 33.76 & \OB{7.97} \\
\midrule
\multirow{4}*{\textbf{Gemma}} & Gemma-2-2B-It & 31.70 & 7.60 & 59.58 & 40.42 & 56.13 & 42.48 & 54.11 & 23.39 & 50.38 & 28.47 & \OA{21.91} \\
~ & Gemma-2-9B-It & 25.30 & 19.00 & 54.86 & 50.92 & 51.45 & 46.90 & 50.06 & 28.82 & 45.42 & 36.41 & \OB{9.01} \\
~ & Gemma-2-27B-It & 40.80 & 31.90 & 67.45 & 62.86 & 71.55 & 66.88 & 63.84 & \textbf{35.02} & 60.91 & 49.16 & \OA{11.75} \\
~ & Gemma-4-31B-it & \textbf{56.20} & \textbf{49.60} & \textbf{77.82} & \textbf{75.33} & \textbf{75.73} & \textbf{70.67} & \textbf{69.66} & 34.77 & \textbf{69.85} & \textbf{57.59} & \OA{12.26} \\
\midrule
\multirow{4}*{\textbf{Phi}} & Phi-3-mini-128k-instruct & 41.60 & 26.00 & 54.46 & 41.08 & 69.53 & 61.82 & 59.17 & 25.54 & 56.19 & 38.61 & \OA{17.58} \\
~ & Phi-3.5-mini-instruct & 32.70 & 19.30 & 59.58 & 43.04 & 65.49 & 55.88 & 60.68 & 26.93 & 54.61 & 36.29 & \OA{18.32} \\
~ & Phi-4-mini-instruct & 47.40 & 36.90 & 62.34 & 51.57 & \textbf{70.04} & \textbf{64.48} & \textbf{62.07} & \textbf{27.43} & \textbf{60.46} & 45.10 & \OA{15.37} \\
~ & Phi-4 & \textbf{52.00} & \textbf{43.50} & \textbf{65.35} & \textbf{60.10} & 67.26 & 60.56 & 53.48 & 23.89 & 59.52 & \textbf{47.01} & \OA{12.51} \\
\midrule
\multirow{2}*{\textbf{Olmo}} & Olmo-3-7B-Instruct & 32.40 & 22.40 & 40.55 & 33.73 & 59.67 & 51.83 & 34.26 & 21.62 & 41.72 & 32.39 & \OB{9.33} \\
~ & Olmo-3.1-32B-Instruct & \textbf{59.00} & \textbf{49.20} & \textbf{73.23} & \textbf{69.16} & \textbf{75.73} & \textbf{69.79} & \textbf{59.17} & \textbf{34.77} & \textbf{66.78} & \textbf{55.73} & \OA{11.05} \\
\bottomrule[1.5pt]
\end{tabular}
\end{threeparttable}}
\vspace{-5mm}
\end{table}

\par
\textbf{Margin faithfulness accuracy.} We also evaluate model behavior under each contextual condition separately: \emph{Answer Accuracy} (AnsAcc) under sufficient context and \emph{Abstention Accuracy} (AbsAcc) under insufficient context:
\begin{equation}
    \mathrm{AnsAcc} = \mathbb{P}(y^+ = \mathrm{C}), \qquad \mathrm{AbsAcc} = \mathbb{P}(y^- = \mathrm{IDK}).
\end{equation}
These two metrics are commonly adopted in existing instance-level evaluation protocols, which primarily focus on answer correctness and abstention accuracy under individual contextual conditions.

\begin{table}[t]
\centering
\caption{\textbf{Evaluation of AnsAcc and AbsAcc on PFaithBench across diverse LLMs.} \textOA{Lightred} indicates that the absolute difference $|\mathrm{AnsAcc} - \mathrm{AbsAcc}|$ is larger than 10\%, while \textOB{lightblue} indicates that the absolute difference is at most 10\%. The best results in each family are highlighted in \textbf{bold}.}
\label{Tble:AnsAcc_AbsAcc}
\vspace{-3mm}
\resizebox{\textwidth}{!}{
\begin{threeparttable}
\begin{tabular}{ll|cc|cc|cc|cc|cc}
    \toprule[1.5pt]
    \multirow{2}*{\textbf{Family}} & \multirow{2}*{\textbf{LLM}} & \multicolumn{2}{c}{\textbf{PFB-MultiHop}} & \multicolumn{2}{c}{\textbf{PFB-SingleHop}} & \multicolumn{2}{c}{\textbf{PFB-ConOri}} & \multicolumn{2}{c}{\textbf{PFB-ConCF}} & \multicolumn{2}{c}{\textbf{Avg.}} \\
    \cmidrule(lr){3-4} \cmidrule(lr){5-6} \cmidrule(lr){7-8} \cmidrule(lr){9-10} \cmidrule(lr){11-12}
    ~ & ~ & \textbf{AnsAcc} $\uparrow$ & \textbf{AbsAcc} $\uparrow$ & \textbf{AnsAcc} $\uparrow$ & \textbf{AbsAcc} $\uparrow$ & \textbf{AnsAcc} $\uparrow$ & \textbf{AbsAcc} $\uparrow$ & \textbf{AnsAcc} $\uparrow$ & \textbf{AbsAcc} $\uparrow$ & \textbf{AnsAcc} $\uparrow$ & \textbf{AbsAcc} $\uparrow$ \\
    \midrule
\multirow{5}*{\textbf{Llama3}} & Llama-3.2-1B-Instruct & 18.60 & 46.10 & 21.92 & 47.24 & 21.49 & 55.75 & 6.19 & 55.63 & \OA{17.05} & \OA{51.18} \\
~ & Llama-3.2-3B-Instruct & 45.90 & 62.60 & 32.28 & \textbf{87.01} & 57.65 & \textbf{80.53} & 18.20 & \textbf{78.89} & \OA{38.51} & \OA{\textbf{77.26}} \\
~ & Llama-3.1-8B-Instruct & 73.80 & 43.40 & 73.36 & 54.59 & 84.45 & 54.11 & 29.58 & 52.97 & \OA{65.30} & \OA{51.27} \\
~ & Llama-3.1-70B-Instruct & \textbf{82.10} & \textbf{62.80} & 86.88 & 79.00 & 85.46 & 71.81 & 30.34 & 69.53 & \OB{71.19} & \OB{70.79} \\
~ & Llama-3.3-70B-Instruct & 81.60 & \textbf{62.80} & \textbf{88.71} & 72.31 & \textbf{86.35} & 67.76 & \textbf{34.51} & 66.88 & \OB{\textbf{72.79}} & \OB{67.44} \\
    \midrule
\multirow{12}*{\textbf{Qwen3}} & Qwen3-0.6B (NoThink) & 36.40 & 43.00 & 59.84 & 50.13 & 80.03 & 55.50 & 43.62 & 58.15 & \OB{54.97} & \OB{51.70} \\
~ & Qwen3-1.7B (NoThink) & 38.20 & 43.00 & 58.79 & 50.52 & 76.61 & 56.89 & 40.20 & 56.64 & \OB{53.45} & \OB{51.76} \\
~ & Qwen3-4B (NoThink) & 56.80 & 67.10 & 73.10 & \textbf{79.66} & 85.59 & 71.81 & 36.92 & 72.06 & \OB{63.10} & \OB{72.66} \\
~ & Qwen3-8B (NoThink) & 55.50 & \textbf{76.80} & 79.27 & 79.13 & 82.68 & \textbf{85.21} & 39.19 & \textbf{85.08} & \OA{64.16} & \OA{\textbf{81.56}} \\
~ & Qwen3-14B (NoThink) & 70.30 & 63.30 & 88.98 & 77.69 & 89.38 & 78.89 & \textbf{51.45} & 75.22 & \OB{75.03} & \OB{73.77} \\
~ & Qwen3-32B (NoThink) & \textbf{82.00} & 58.80 & \textbf{92.91} & 64.96 & \textbf{90.14} & 72.31 & 47.79 & 68.65 & \OA{\textbf{78.21}} & \OA{66.18} \\
    \cmidrule(lr){2-12}
~ & Qwen3-0.6B (Think) & 61.90 & 15.10 & 78.87 & 11.94 & 88.24 & 19.22 & 48.42 & 16.56 & \OA{69.36} & \OA{15.70} \\
~ & Qwen3-1.7B (Think) & 63.00 & 14.40 & 78.35 & 10.89 & 88.12 & 18.20 & 48.67 & 17.83 & \OA{69.53} & \OA{15.33} \\
~ & Qwen3-4B (Think) & 81.20 & 53.70 & 89.11 & 58.66 & 87.61 & \textbf{59.54} & \textbf{50.19} & \textbf{58.79} & \OA{77.03} & \OA{57.67} \\
~ & Qwen3-8B (Think) & 81.00 & 53.30 & 88.71 & \textbf{64.70} & 87.48 & 59.29 & 48.04 & 57.77 & \OA{76.31} & \OA{\textbf{58.77}} \\
~ & Qwen3-14B (Think) & 82.90 & \textbf{54.20} & 91.34 & 55.51 & 86.85 & 50.95 & 49.18 & 43.11 & \OA{77.57} & \OA{50.94} \\
~ & Qwen3-32B (Think) & \textbf{85.90} & 48.00 & \textbf{93.96} & 52.10 & \textbf{90.39} & 39.70 & 48.17 & 34.64 & \OA{\textbf{79.61}} & \OA{43.61} \\
    \midrule
\multirow{10}*{\textbf{Qwen3.5}} & Qwen3.5-0.8B (NoThink) & 59.30 & 46.20 & 54.46 & 63.78 & 75.60 & 73.83 & 38.18 & 75.09 & \OB{56.89} & \OB{64.73} \\
~ & Qwen3.5-2B (NoThink) & 71.20 & 39.50 & 78.87 & 53.54 & 81.42 & 68.77 & 42.86 & 64.98 & \OA{68.59} & \OA{56.70} \\
~ & Qwen3.5-4B (NoThink) & 81.50 & 48.00 & 88.71 & 68.64 & 89.13 & 70.54 & 46.78 & 71.55 & \OA{76.53} & \OA{64.68} \\
~ & Qwen3.5-9B (NoThink) & 82.90 & \textbf{53.00} & 93.96 & 69.16 & 87.86 & 73.96 & \textbf{53.35} & 75.35 & \OA{\textbf{79.52}} & \OA{67.87} \\
~ & Qwen3.5-27B (NoThink) & \textbf{84.10} & 51.20 & \textbf{94.23} & \textbf{72.18} & \textbf{89.38} & \textbf{78.00} & 47.28 & \textbf{77.12} & \OB{78.75} & \OB{\textbf{69.62}} \\
    \cmidrule(lr){2-12}
~ & Qwen3.5-0.8B (Think) & 64.70 & 23.20 & 78.22 & 37.53 & 80.78 & 44.63 & 39.44 & 46.14 & \OA{65.79} & \OA{37.88} \\
~ & Qwen3.5-2B (Think) & 80.10 & 36.20 & 88.32 & 62.20 & 85.34 & 66.50 & 45.64 & 65.99 & \OA{74.85} & \OA{57.72} \\
~ & Qwen3.5-4B (Think) & 83.60 & 25.50 & \textbf{92.91} & 44.75 & \textbf{87.99} & 50.06 & 48.55 & 49.18 & \OA{78.26} & \OA{42.37} \\
~ & Qwen3.5-9B (Think) & 84.00 & \textbf{67.80} & 90.29 & \textbf{81.89} & 87.10 & \textbf{91.53} & \textbf{52.34} & \textbf{89.51} & \OB{\textbf{78.43}} & \OB{\textbf{82.68}} \\
~ & Qwen3.5-27B (Think) & \textbf{84.20} & 19.90 & 92.39 & 49.48 & 86.60 & 59.92 & 47.16 & 56.64 & \OA{77.59} & \OA{46.48} \\
    \midrule
\multirow{2}*{\textbf{Qwen3.6}} & Qwen3.6-27B (NoThink) & 83.90 & \textbf{58.70} & 94.36 & \textbf{73.75} & 88.37 & \textbf{77.62} & 47.41 & \textbf{78.51} & \OB{78.51} & \OB{\textbf{72.15}} \\
~ & Qwen3.6-27B (Think) & \textbf{86.10} & 20.90 & \textbf{95.80} & 45.54 & \textbf{89.63} & 54.74 & \textbf{54.24} & 52.59 & \OA{\textbf{81.44}} & \OA{43.44} \\
    \midrule
\multirow{4}*{\textbf{Gemma}} & Gemma-2-2B-It & 18.90 & 56.00 & 58.66 & 70.47 & 69.28 & 61.44 & 34.01 & 63.08 & \OA{45.21} & \OA{62.75} \\
~ & Gemma-2-9B-It & 75.00 & 29.10 & 88.58 & 57.48 & \textbf{89.00} & 53.98 & \textbf{48.67} & 53.73 & \OA{75.31} & \OA{48.57} \\
~ & Gemma-2-27B-It & 76.20 & 46.60 & 88.32 & 69.69 & 88.50 & 76.23 & 43.99 & 70.92 & \OB{74.25} & \OB{65.86} \\
~ & Gemma-4-31B-it & \textbf{84.50} & \textbf{61.20} & \textbf{94.09} & \textbf{80.05} & 85.46 & \textbf{84.32} & 42.10 & \textbf{82.43} & \OB{\textbf{76.54}} & \OB{\textbf{77.00}} \\
    \midrule
\multirow{4}*{\textbf{Phi}} & Phi-3-mini-128k-instruct & 49.00 & \textbf{65.20} & 54.33 & \textbf{82.15} & 77.37 & \textbf{82.43} & 28.19 & \textbf{82.05} & \OA{52.22} & \OA{\textbf{77.96}} \\
~ & Phi-3.5-mini-instruct & 56.10 & 44.70 & 60.37 & 74.15 & 81.04 & 69.91 & 32.62 & 70.42 & \OB{57.53} & \OB{64.79} \\
~ & Phi-4-mini-instruct & 65.50 & 61.30 & 69.55 & 77.43 & \textbf{85.34} & 76.74 & \textbf{35.02} & 74.08 & \OB{63.85} & \OB{72.39} \\
~ & Phi-4 & \textbf{79.70} & 57.70 & \textbf{85.83} & 69.82 & 83.19 & 75.22 & 27.31 & 72.82 & \OB{\textbf{69.00}} & \OB{68.89} \\
    \midrule
\multirow{2}*{\textbf{Olmo}} & Olmo-3-7B-Instruct & 30.80 & \textbf{80.40} & 37.40 & \textbf{93.31} & 59.29 & \textbf{89.63} & 23.89 & \textbf{87.74} & \OA{37.85} & \OA{\textbf{87.77}} \\
~ & Olmo-3.1-32B-Instruct & \textbf{66.90} & 78.10 & \textbf{79.66} & 86.75 & \textbf{82.68} & 85.97 & \textbf{40.58} & 81.54 & \OA{\textbf{67.46}} & \OA{83.09} \\
    \bottomrule[1.5pt]
\end{tabular}
\end{threeparttable}}
\vspace{-5mm}
\end{table}

\subsection{PFaithBench Construction}
\label{Sec:PFaithBench_Contruct}
To systematically evaluate pairwise faithfulness, we construct PFaithBench $\mathcal{D}=\{(q,c^+,c^-,a)\}$, where each example $(q,c^+,c^-,a)$ pairs a sufficient and an insufficient context for the same question. As illustrated in Fig.~\ref{Fig:overview}, we collect sufficient-context QA examples from three types of data sources: multi-hop, single-hop, and counterfactual QA. We then construct $c^-$ by minimally substituting, removing, or blurring key information in $c^+$, followed by manual verification to ensure its quality.

\textbf{Multi-hop factual QA.}
We collect $(q,c^+,a)$ from HotpotQA~\citep{yang2018hotpotqa} and 2WikiMultihopQA~\citep{ho20202wiki}, where answering requires multiple supporting facts. Both datasets provide annotated supporting facts, allowing us to construct $c^-$ by replacing one key supporting fact with an irrelevant fact, breaking the evidence dependency, resulting in the PFB-MultiHop subset.

\textbf{Single-hop factual QA.}
We collect $(q,c^+,a)$ from SQuAD~\citep{rajpurkar2016squad}. The insufficient context $c^-$ is constructed using LLM rewriting to remove or blur key answer-supporting information while keeping the remaining context largely unchanged, resulting in the PFB-SingleHop subset.

\textbf{Counterfactual QA.} 
Counterfactual QA refers to questions that require answering based on contexts that contradict real-world knowledge~\citep{mingfaitheval}. We collect $(q, c^+_{\mathrm{ori}}, a_{\mathrm{ori}}, c^+_{\mathrm{cf}}, a_{\mathrm{cf}})$ from ConFiQA~\citep{bi2025contextDPO}, which provides paired original and counterfactual contexts for the same question $q$. We treat both $c^+_{\mathrm{ori}}$ and $c^+_{\mathrm{cf}}$ as sufficient contexts and construct their corresponding insufficient contexts by removing or blurring only the key answer-supporting information while preserving the remaining content, resulting in the PFB-ConOri and PFB-ConCF subsets.

\par
For each QA type, we begin with 1,000 examples of the form $(q,c^+,a)$. For multi-hop QA, annotated supporting facts allow us to construct insufficient contexts by replacing a key fact with an irrelevant one. For single-hop and counterfactual QA, two postgraduate students manually verify the LLM-rewritten insufficient contexts, yielding 762 and 791 valid examples, respectively. The higher retention rates $76.2\%$ and $79.1\%$ demonstrate the effectiveness of our construction pipeline. PFaithBench thus comprises four evaluation subsets: PFB-MultiHop, PFB-SingleHop, PFB-ConOri, and PFB-ConCF, with 1,000, 762, 791 and 791 examples, respectively. Additionally, we use an LLM judge to classify model responses as $\{\mathrm{IDK},\mathrm{C},\mathrm{W}\}$ for metric computation. More details refer to Appendix~\ref{App:details_about_PFaithBench}.

\begin{table}[t]
\centering
\vspace{-2mm}
\caption{\textbf{Impact of context coupling on faithfulness training.} The best value of each metric is \textbf{bold}. Among the three training settings, \textOB{lightblue} marks both AnsAcc and AbsAcc for the smallest $|\mathrm{AnsAcc}-\mathrm{AbsAcc}|$, and marks both OverAns and OverAbs for the smallest $|\mathrm{OverAns}-\mathrm{OverAbs}|$.}
\label{Tble:MismatchedContext}
\vspace{-3mm}
\resizebox{\textwidth}{!}{
\begin{threeparttable}
\begin{tabular}{ll|cc|cc|cc|cc|cc|cc}
    \toprule[1.5pt]
    \multirow{2}*{\textbf{LLM}} & \multirow{2}*{\textbf{Setting}} & \multicolumn{6}{c}{\textbf{PFB-MultiHop}} & \multicolumn{6}{c}{\textbf{PFB-SingleHop}} \\
    \cmidrule(lr){3-8} \cmidrule(lr){9-14}
    ~ & ~ & \textbf{DFAcc} $\uparrow$ & \textbf{SFAcc} $\uparrow$ & \textbf{AnsAcc} $\uparrow$ & \textbf{AbsAcc} $\uparrow$ & \textbf{OverAns} $\downarrow$ & \textbf{OverAbs} $\downarrow$ & \textbf{DFAcc} $\uparrow$ & \textbf{SFAcc} $\uparrow$ & \textbf{AnsAcc} $\uparrow$ & \textbf{AbsAcc} $\uparrow$ & \textbf{OverAns} $\downarrow$ & \textbf{OverAbs} $\downarrow$ \\
    \midrule
    \multirow{4}*{\textbf{Llama-3.1-8B-Instruct}} & Base Model & 37.70 & 30.00 & 73.80 & 43.40 & 56.60 & 6.50 & 49.87 & 39.50 & 73.36 & 54.59 & 45.41 & 6.04 \\
    \cmidrule(lr){2-14}
    ~ & MHSuff-SHInsuff & 34.20 & 27.40 & \textbf{81.80} & 36.60 & 63.40 & \textbf{2.60} & 37.14 & 31.50 & 32.28 & \textbf{98.56} & \textbf{1.44} & 61.68 \\
    ~ & SHSuff-MHInsuff & 2.70 & 0.00 & 0.00 & \textbf{96.70} & \textbf{3.30} & 95.60 & 16.01 & 14.70 & \textbf{91.60} & 16.80 & 83.20 & \textbf{1.05} \\
    ~ & \OB{\textbf{PairContext}} & \textbf{71.50} & \textbf{61.00} & \OB{76.40} & \OB{81.90} & \OB{18.10} & \OB{10.60} & \textbf{79.13} & \textbf{75.33} & \OB{86.22} & \OB{86.35} & \OB{13.65} & \OB{7.74} \\
    \midrule
    \multirow{4}*{\textbf{Llama-3.2-3B-Instruct}} & Base Model & 42.20 & 26.00 & 45.90 & 62.60 & 37.40 & 23.90 & 46.19 & 25.85 & 32.28 & 87.01 & 12.99 & 43.18 \\
    \cmidrule(lr){2-14}
    ~ & MHSuff-SHInsuff & 33.80 & 26.00 & \textbf{73.00} & 38.20 & 61.80 & \textbf{6.30} & 26.90 & 23.10 & 24.93 & \textbf{97.51} & \textbf{2.49} & 70.87 \\
    ~ & SHSuff-MHInsuff & 45.00 & 10.90 & 18.70 & 79.60 & 20.40 & 37.00 & 36.35 & 33.07 & \textbf{82.55} & 39.90 & 60.10 & \textbf{4.20} \\
    ~ & \OB{\textbf{PairContext}} & \textbf{68.20} & \textbf{53.40} & \OB{69.40} & \OB{\textbf{80.00}} & \OB{\textbf{20.00}} & \OB{12.50} & \textbf{67.45} & \textbf{62.86} & \OB{79.27} & \OB{79.40} & \OB{20.60} & \OB{12.47} \\
    \midrule
    \multirow{4}*{\textbf{Qwen3-4B (NoThink)}} & Base Model & 44.10 & 32.90 & 56.80 & 67.10 & 32.90 & 25.80 & 63.12 & 56.56 & 73.10 & 79.66 & 20.34 & 18.24 \\
    \cmidrule(lr){2-14}
    ~ & MHSuff-SHInsuff & 56.10 & 45.50 & \textbf{74.20} & 64.00 & 36.00 & \textbf{8.70} & 67.72 & 56.30 & 61.02 & \textbf{93.44} & \textbf{6.56} & 26.51 \\
    ~ & SHSuff-MHInsuff & 58.40 & 47.30 & \OB{73.30} & \OB{67.10} & 32.90 & 9.60 & 69.42 & 63.78 & \textbf{86.48} & 73.62 & 26.38 & \textbf{5.25} \\
    ~ & \OB{\textbf{PairContext}} & \textbf{68.80} & \textbf{60.50} & 74.10 & \textbf{84.20} & \OB{\textbf{15.80}} & \OB{16.00} & \textbf{77.95} & \textbf{74.15} & \OB{81.63} & \OB{90.29} & \OB{9.71} & \OB{12.60} \\
    \midrule
    \multirow{4}*{\textbf{Qwen3-1.7B (NoThink)}} & Base Model & 28.40 & 11.40 & 38.20 & 43.00 & 57.00 & 17.90 & 41.47 & 26.90 & 58.79 & 50.52 & 49.48 & 10.76 \\
    \cmidrule(lr){2-14}
    ~ & MHSuff-SHInsuff & 24.10 & 14.90 & \textbf{61.10} & 27.20 & 72.80 & \textbf{4.20} & 43.57 & 19.16 & 21.26 & \textbf{93.96} & \textbf{6.04} & 51.18 \\
    ~ & SHSuff-MHInsuff & 32.20 & 13.90 & 26.30 & 75.60 & 24.40 & 46.70 & 38.58 & 31.50 & \textbf{74.54} & 42.65 & 57.35 & \textbf{5.12} \\
    ~ & \OB{\textbf{PairContext}} & \textbf{61.50} & \textbf{43.70} & \OB{58.50} & \OB{\textbf{77.20}} & \OB{\textbf{22.80}} & \OB{17.80} & \textbf{58.01} & \textbf{50.52} & \OB{65.88} & \OB{79.53} & \OB{20.47} & \OB{23.62} \\
    \midrule
    \multirow{2}*{\textbf{LLM}} & \multirow{2}*{\textbf{Setting}} & \multicolumn{6}{c}{\textbf{PFB-ConOri}} & \multicolumn{6}{c}{\textbf{PFB-ConCF}} \\
    \cmidrule(lr){3-8} \cmidrule(lr){9-14}
    ~ & ~ & \textbf{DFAcc} $\uparrow$ & \textbf{SFAcc} $\uparrow$ & \textbf{AnsAcc} $\uparrow$ & \textbf{AbsAcc} $\uparrow$ & \textbf{OverAns} $\downarrow$ & \textbf{OverAbs} $\downarrow$ & \textbf{DFAcc} $\uparrow$ & \textbf{SFAcc} $\uparrow$ & \textbf{AnsAcc} $\uparrow$ & \textbf{AbsAcc} $\uparrow$ & \textbf{OverAns} $\downarrow$ & \textbf{OverAbs} $\downarrow$ \\
    \midrule
    \multirow{4}*{\textbf{Llama-3.1-8B-Instruct}} & Base Model & 50.95 & 45.13 & 84.45 & 54.11 & 45.89 & 3.79 & 42.10 & 19.09 & 29.58 & 52.97 & 47.03 & 18.46 \\
    \cmidrule(lr){2-14}
    ~ & MHSuff-SHInsuff & 66.75 & 61.44 & 66.62 & \textbf{94.06} & \textbf{5.94} & 27.69 & 42.98 & 24.78 & 25.92 & \textbf{94.94} & \textbf{5.06} & 53.86 \\
    ~ & SHSuff-MHInsuff & 15.80 & 14.29 & \textbf{91.40} & 16.81 & 83.19 & \textbf{1.14} & 12.39 & 7.59 & \OB{\textbf{52.72}} & \OB{14.79} & 85.21 & \textbf{3.79} \\
    ~ & \OB{\textbf{PairContext}} & \textbf{80.40} & \textbf{75.98} & \OB{85.84} & \OB{89.00} & \OB{11.00} & \OB{9.10} & \textbf{70.16} & \textbf{42.35} & 47.03 & 87.74 & \OB{12.26} & \OB{19.22} \\
    \midrule
    \multirow{4}*{\textbf{Llama-3.2-3B-Instruct}} & Base Model & 59.29 & 43.87 & 57.65 & 80.53 & 19.47 & 23.51 & 47.91 & 15.93 & 18.20 & 78.89 & 21.11 & 37.42 \\
    \cmidrule(lr){2-14}
    ~ & MHSuff-SHInsuff & 65.36 & 52.72 & 57.77 & \textbf{93.30} & \textbf{6.70} & 28.45 & 48.17 & 19.97 & 20.73 & \textbf{95.07} & \textbf{4.93} & 49.30 \\
    ~ & SHSuff-MHInsuff & 52.84 & 49.18 & \textbf{88.24} & 57.14 & 42.86 & \textbf{4.80} & 39.19 & 23.14 & \OB{38.56} & \OB{53.35} & 46.65 & \textbf{21.74} \\
    ~ & \OB{\textbf{PairContext}} & \textbf{78.00} & \textbf{73.32} & \OB{85.71} & \OB{86.22} & \OB{13.78} & \OB{8.22} & \textbf{63.46} & \textbf{40.83} & \textbf{46.40} & 84.70 & \OB{15.30} & \OB{24.40} \\
    \midrule
    \multirow{4}*{\textbf{Qwen3-4B (NoThink)}} & Base Model & 63.72 & 59.29 & 85.59 & 71.81 & 28.19 & 8.22 & 55.12 & 29.58 & 36.92 & 72.06 & 27.94 & 22.12 \\
    \cmidrule(lr){2-14}
    ~ & MHSuff-SHInsuff & 76.36 & 65.99 & 76.36 & 86.73 & \OB{13.27} & \OB{11.38} & 63.08 & 32.49 & 35.90 & 87.61 & 12.39 & 27.18 \\
    ~ & SHSuff-MHInsuff & 70.80 & 65.87 & \OB{\textbf{86.09}} & \OB{77.75} & 22.25 & \textbf{6.95} & 63.97 & 34.39 & \OB{\textbf{41.09}} & \OB{77.37} & \OB{22.63} & \OB{\textbf{15.68}} \\
    ~ & \OB{\textbf{PairContext}} & \textbf{80.66} & \textbf{75.60} & 81.29 & \textbf{94.18} & \textbf{5.82} & 13.65 & \textbf{68.65} & \textbf{39.70} & 40.71 & \textbf{94.56} & \textbf{5.44} & 28.07 \\
    \midrule
    \multirow{4}*{\textbf{Qwen3-1.7B (NoThink)}} & Base Model & 52.47 & 41.97 & 76.61 & 56.89 & 43.11 & \textbf{5.69} & 49.30 & 24.78 & 40.20 & 56.64 & 43.36 & \textbf{8.98} \\
    \cmidrule(lr){2-14}
    ~ & MHSuff-SHInsuff & 67.89 & 44.75 & 49.94 & \textbf{91.40} & \textbf{8.60} & 23.89 & 61.44 & 20.86 & 22.63 & \textbf{91.15} & \textbf{8.85} & 30.47 \\
    ~ & SHSuff-MHInsuff & 50.82 & 46.90 & \textbf{84.20} & 57.27 & 42.73 & 7.21 & 50.06 & 27.69 & \OB{\textbf{47.16}} & \OB{56.01} & 43.99 & \textbf{8.98} \\
    ~ & \OB{\textbf{PairContext}} & \textbf{68.39} & \textbf{62.83} & \OB{79.39} & \OB{81.42} & \OB{18.58} & \OB{13.40} & \textbf{61.69} & \textbf{34.26} & 41.72 & 80.66 & \OB{19.34} & \OB{20.99} \\
    \bottomrule[1.5pt]
\end{tabular}
\end{threeparttable}}
\vspace{-5mm}
\end{table}

\subsection{Evaluation Results}
A wide range of existing LLMs is taken into the evaluation, covering seven families with thirty-nine models across parameters from 0.6 billion to 70 billion. All models are explicitly prompted to answer solely based on the provided context. When the context is sufficient, the model is required to provide an answer along with its reasoning process; otherwise, the model should acknowledge the lack of information and respond ``I don't know''. The detailed evaluation setup is provided in Appendix~\ref{App:EvaluationDetails}.

\textbf{Most existing models exhibit a strong bias toward over-answering.} As shown in Table~\ref{Tble:OverAns_OverAbs}, the majority of models $35/39$ demonstrate substantially higher OverAns than OverAbs, and only a few models, such as Olmo-3-7B, show the over-abstention-dominant behavior. indicating a tendency that models produce answers even when the context is insufficient. One possible reason is existing outcome-based post-training, such as reinforcement learning with verifiable rewards (RLVR)~\citep{yang2026scaling}. Such methods primarily reward correct answers and penalize incorrect ones, while often treating appropriate abstention as an incorrect response. This may encourage models to attempt an answer even when the context does not include sufficient information~\citep{yao2025reasoning}.

\textbf{Over-answering is closely associated with internal knowledge leakage.} 
Figure~\ref{Fig:Leakage} shows that all models exhibit some degree of internal knowledge leakage: when the provided context is insufficient, they draw on their internal knowledge to produce an answer. This may partly explain their tendency to over-answer. Within most model families, leakage also increases with model size, suggesting that larger models are more likely to rely on internal knowledge when context is insufficient. Although this behavior may make models appear more helpful, it also increases the risk of unsupported answers in information-sensitive settings, particularly when their internal knowledge is outdated or incorrect.

\textbf{Models exhibit imbalanced and asymmetric context-sensitive behavior.}
Tables~\ref{Tble:AnsAcc_AbsAcc} and~\ref{Tble:DFAcc_SFAcc} report the marginal AnsAcc and AbsAcc and joint DFAcc and SFAcc metrics, respectively. Many models show a substantial gap between AnsAcc and AbsAcc, indicating an imbalance between answering and abstention. Their joint performance is also weaker than their marginal performance, suggesting that strong accuracy under individual context conditions may not translate into pairwise faithful behavior.

\begin{figure}[t]
  \centering
  \vspace{2mm}
  \includegraphics[width=0.325\textwidth]{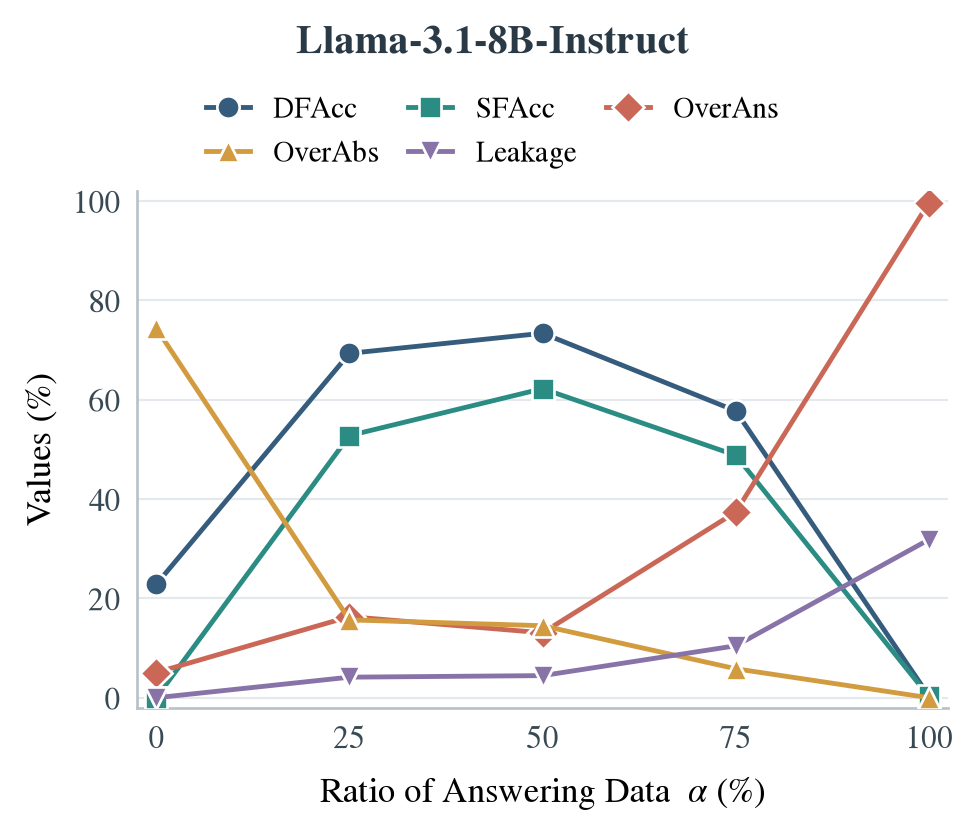}
  \includegraphics[width=0.325\textwidth]{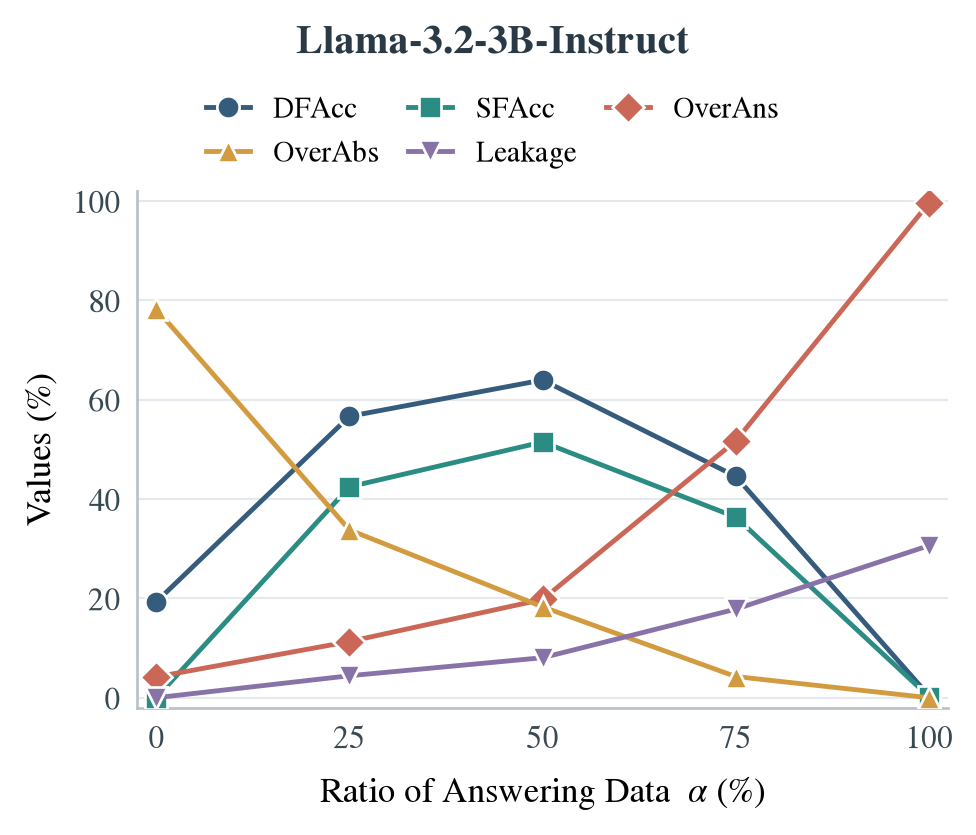}
  \includegraphics[width=0.325\textwidth]{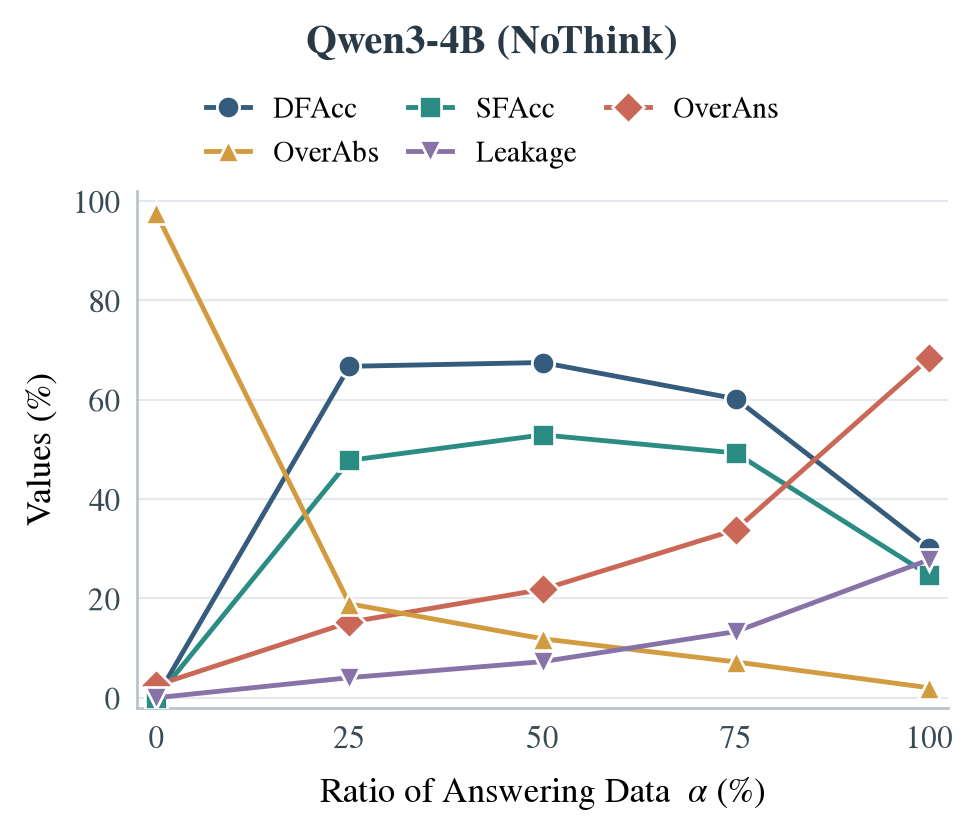}
  \vspace{-3mm}
  \caption{\textbf{Impact of answering–abstention data composition (asymmetric reward).} Faithfulness training under different answering data ratios $\alpha \in \{0,25,50,75,100\}$ across three LLMs. This is the average value across four subsets in PFaithBench. More results are provided in Appendix~\ref{App:Symmetric_reward}.}
  \vspace{-5mm}
  \label{Fig:data_composition_asymmetric}
\end{figure}

\section{Data-centric Analyses for Faithfulness Training}
\label{Sec:DataCentric_Analysis}
PFaithBench pairs sufficient and insufficient contexts $(c^+,c^-)$ for each question, providing a natural setting to examine how data composition affects faithfulness. In this section, we investigate two factors: the coupling between sufficient and insufficient contexts, and their relative proportions.

\subsection{Faithfulness Training Scheme}
\label{Sec:Asymmetric_reward}
Existing faithfulness training has explored supervised fine-tuning (SFT)~\citep{zhang2024RTuning}, direct preference optimization (DPO)~\citep{bi2025contextDPO}, and reinforcement learning (RL)~\citep{gui2026FaithRL,ren2025knowrl}, primarily focusing on response correctness under given contexts. In our pairwise faithfulness setting, model responses are explicitly categorized into $\{\mathrm{C}, \mathrm{W}, \mathrm{IDK}\}$, enabling explicit rewards for answering and abstention. We therefore adopt GRPO~\citep{shao2024deepseekmath} as the training algorithm, and define rewards that penalize both over-answering and over-abstention:
\begin{equation}
r(a,y^-)=
\begin{cases}
+1 & \text{if } y^-=\mathrm{IDK},\\
-1 & \text{if } y^-\in\{\mathrm{C},\mathrm{W}\},
\end{cases}
\quad
r_{\mathrm{asym}}(a,y^+)=
\begin{cases}
+1 & \text{if } y^+=\mathrm{C},\\
0 & \text{if } y^+=\mathrm{W},\\
-1 & \text{if } y^+=\mathrm{IDK},
\end{cases}
\end{equation}
Under sufficient context ($c^+$), this reward favors answering over abstaining, even when the answer is incorrect, while assigning the highest reward to a correct answer. Under insufficient context ($c^-$), it rewards abstention and penalizes answering. The subsequent analyses are conducted on this training scheme and training data is constructed using the same pipeline as in PFaithBench. We also examine a symmetric reward that penalizes incorrect answers and abstention equally under $c^+$ in Appendix~\ref{App:Symmetric_reward}.

\subsection{Confounding Effects from Mismatched Context Coupling}
\label{Sec:Mismatched Context Coupling}
We investigate how the coupling between sufficient and insufficient contexts affects faithfulness training, particularly when these contexts are drawn from different data sources. The question is whether models learn to rely on contextual sufficiency or instead exploit dataset-specific shortcuts.
\begin{itemize}[leftmargin=0.3cm]
    \item \textbf{Mismatched context coupling:} Sufficient and insufficient examples come from different QA sources, $(q_A,c^+)\sim\mathcal{D}_{\mathrm A}$ and $(q_B,c^-)\sim\mathcal{D}_{\mathrm B}$. Thus, each question appears with only one context type, making sufficiency correlated with data source. We consider two  settings: MHSuff-SHInsuff (multi-hop sufficient, single-hop insufficient) and SHSuff-MHInsuff (the reverse).
    \item Both contexts are observed for each question, $(q,c^+,c^-)\sim\mathcal{D}$, preserving their pairing. We sample these pairs from a balanced pool of multi-hop and single-hop QA data, denoted PairContext.
\end{itemize}
The experiments are conducted across four models of varying scales: Llama-3.1-8B-Instruct, Llama-3.2-3B-Instruct, and Qwen3-4B/1.7B. Table~\ref{Tble:MismatchedContext} reports the results across four subsets of PFaithBench. 


\textbf{Mismatched context coupling may induce data-specific shortcuts.} As shown in Table~\ref{Tble:MismatchedContext}, mismatched training produces behavior aligned with data source. For example, Llama-3.1-8B-Instruct trained with MHSuff-SHInsuff shows high OverAns on PFB-MultiHop (63.40) and high OverAbs on PFB-SingleHop (61.68). Reversing the sources leads to high OverAbs on PFB-MultiHop (95.60) and high OverAns on PFB-SingleHop (83.20). This pattern suggests that models may use source-specific cues to decide whether to answer. PairContext achieves the best pairwise faithfulness in most cases, indicating the importance of preserving context pairs during faithfulness training.

\textbf{Mismatched context coupling may exacerbate the imbalance between answering and abstention.} From the Table~\ref{Tble:MismatchedContext}, we observe that mismatched context coupling can produce extreme imbalances between AnsAcc and AbsAcc, with one becoming very high while the other remains very low. For example, under MHSuff-SHInsuff, Llama-3.1-8B-Instruct achieves high AnsAcc but low AbsAcc on PFB-MultiHop (81.80 vs. 36.60), while showing the reverse on PFB-SingleHop (32.28 vs. 98.56). This suggests that mismatched training may induce a source-conditioned answer-abstention bias.

\begin{table}[t]
\centering
\vspace{-2mm}
\caption{\textbf{Impact of faithfulness training on general model capabilities.}}
\label{Tble:General_capability}
\vspace{-3mm}
\resizebox{\textwidth}{!}{
\begin{threeparttable}
\begin{tabular}{l|ccccccc|l|ccccccc}
    \toprule[1.5pt]
    \textbf{Setting} & \textbf{AMC} & \textbf{AIME24} & \textbf{AIME25} & \textbf{MATH500} & \textbf{MMLU-Pro} & \textbf{GPQA-D} & \textbf{IFEval} & \textbf{Setting} & \textbf{AMC} & \textbf{AIME24} & \textbf{AIME25} & \textbf{MATH500} & \textbf{MMLU-Pro} & \textbf{GPQA-D} & \textbf{IFEval} \\
    \midrule
    \multicolumn{8}{c|}{\textit{\textbf{Llama-3.1-8B-Instruct}}} & $\alpha=0$ & 16.42 & 3.02 & 0.42 & 31.80 & 34.21 & 26.26 & 78.82 \\
    \hhline{-|-------|~|~~~~~~~}
    Base Model & 14.76 & 2.19 & 0.63 & 33.40 & 36.10 & 29.29 & 76.86 & $\alpha=25$ & 17.32 & 3.33 & 0.21 & 34.60 & 36.43 & 30.81 & 77.42 \\
    MHSuff-SHInsuff & 19.28	& 3.96 & 0.52 & 38.60 & 35.61 & 25.76 & 80.34 & $\alpha=50$ & 	18.37 & 3.33 & 0.52 & 34.60 & 36.18 & 25.25 & 78.87 \\
    SHSuff-MHInsuff & 18.52 & 2.71 & 0.31 & 36.20 & 35.46 & 35.35 & 79.53 & $\alpha=75$ & 17.62 & 3.23 & 0.31 & 37.60 & 35.85 & 27.78 & 78.53 \\
    PairContext & 19.43 & 4.17 & 0.31 & 38.00 & 36.58 & 30.30 & 77.81 & $\alpha=100$ & 18.52 & 4.17 & 0.31 & 36.80 & 36.58 & 27.78 & 78.39 \\
    \specialrule{1pt}{0.4ex}{0.4ex}
    \multicolumn{8}{c|}{\textit{\textbf{Llama-3.2-3B-Instruct}}} & $\alpha=0$ & 15.06 & 5.42 & 0.10 & 34.60 & 28.99 & 26.77 & 74.14 \\
    \hhline{-|-------|~|~~~~~~~}
    Base Model & 17.17 & 4.06 & 0.21 & 35.00 & 29.73 & 31.82 & 72.97 & $\alpha=25$ & 15.06 & 5.63 & 0.42 & 35.00 & 29.31 & 28.79 & 74.79 \\
    MHSuff-SHInsuff & 16.87 & 5.94 & 0.31 & 36.40 & 29.93 & 26.77 & 73.61 & $\alpha=50$ & 	18.07 & 4.90 & 0.42 & 36.20 & 29.34 & 24.24 & 72.53 \\
    SHSuff-MHInsuff & 17.77 & 4.69 & 0.83 & 37.60 & 29.13 & 27.27 & 71.72 & $\alpha=75$ & 17.77 & 4.17 & 0.21 & 36.60 & 29.73 & 28.79 & 74.28 \\
    PairContext & 17.77 & 5.42 & 70.42 & 35.00 & 29.08 & 25.25 & 74.83 & $\alpha=100$ & 18.98 & 5.21 & 0.42 & 34.60 & 30.27 & 26.77 & 73.49 \\
    \specialrule{1pt}{0.4ex}{0.4ex}
    \multicolumn{8}{c|}{\textit{\textbf{Qwen3-4B (NoThink)}}} & $\alpha=0$ & 58.43 & 21.88 & 17.40 & 80.80 & 30.94 & 47.98 & 84.59 \\
    \hhline{-|-------|~|~~~~~~~}
    Base Model & 59.04 & 21.98 & 18.23 & 79.80 & 34.88 & 44.44 & 82.79 & $\alpha=25$ & 57.83 & 22.29 & 18.02 & 80.00 & 32.75 & 46.46 & 84.16 \\
    MHSuff-SHInsuff & 59.19 & 22.40 & 17.50 & 79.40 & 34.72 & 42.93 & 84.38 & $\alpha=50$ & 58.43 & 22.19 & 18.54 & 81.00 & 32.26 & 43.94 & 83.42 \\
    SHSuff-MHInsuff & 57.98 & 23.44 & 18.54 & 81.00 & 32.77 & 44.44 & 84.06 & $\alpha=75$ & 57.23 & 22.08 & 18.23 & 81.40 & 35.41 & 42.42 & 84.42 \\
    PairContext & 57.23 & 21.46 & 17.81 & 81.40 & 31.02 & 41.41 & 83.64 & $\alpha=100$ & 57.38 & 22.60 & 16.98 & 83.80 & 35.66 & 41.41 & 84.57 \\
    \bottomrule[1.5pt]
\end{tabular}
\end{threeparttable}}
\vspace{-5mm}
\end{table}

\subsection{Impact of Answering-Abstention Data Proportion}
Beyond context coupling, we further investigate how the proportion of sufficient-context (answering) and insufficient-context (abstaining) data affects faithfulness training. Specifically, we vary the percentage of answering data as $\alpha \in \{0, 25, 50, 75, 100\}$ from abstention-only ($\alpha=0$) to answering-only ($\alpha=100$). We conduct training under these different data compositions across three models: Llama-3.1-8B-Instruct, Llama-3.2-3B-Instruct, and Qwen3-4B (NoThink).

\textbf{Data composition shapes the balance between answering and abstention.}
Figure~\ref{Fig:data_composition_asymmetric} shows that increasing $\alpha$ generally raises OverAns and Leakage while reducing OverAbs. At $\alpha=0$, abstention dominates and SFAcc falls to zero; at $\alpha=100$, over-answering becomes severe, reaching 100\% for both Llama models. Across all three models, DFAcc and SFAcc peak near the balanced setting ($\alpha=50$). These results suggest that a balanced mix of sufficient- and insufficient-context data helps faithfulness training improve models’ ability to switch between answering and abstaining.

\par
\textbf{Faithfulness training does not affect general model capabilities.} Table~\ref{Tble:General_capability} further demonstrate whether faithfulness training affects models' general capabilities. Specifically, we evaluate the base and faithfulness-trained models across diverse benchmarks, covering mathematical reasoning AMC~\citep{yang2024qwen2p5math}, AIME24~\citep{aime24}, AIME25~\citep{aime25}, and MATH500~\citep{lightman2024math500}; multi-domain knowledge MMLU-Pro~\citep{wang2024mmlu} and GPQA-Diamond~\citep{rein2023gpqa}; and instruction-following IFEval~\citep{zhou2023ifeval}. We observe that faithfulness-trained models retain performance comparable to their untrained base model. These results suggest that the changes in pairwise faithfulness need not come at the expense of general capabilities, supporting the study of faithfulness as a distinct training capability.

\vspace{-2mm}
\section{Conclusion}
In this work, we revisit faithfulness in LLMs from a pairwise context-sensitive perspective and introduce PFaithBench, a benchmark that evaluates whether models can adapt their behavior for the same question under sufficient and insufficient contexts. Unlike instance-level evaluation, our framework explicitly measures the ability of models to switch between answering and abstaining based on contextual sufficiency. Our evaluation identifies over-answering and over-abstention as the main faithfulness errors, with existing models biased toward over-answering, which is closely associated with internal knowledge leakage. Our data-centric analyses show that context coupling and the balance of answering and abstention data matter for faithfulness training.

\bibliography{iclr2027_conference}
\bibliographystyle{iclr2027_conference}

\newpage
\appendix

\section{More Details of PFaithBench}
\label{App:details_about_PFaithBench}

\subsection{Details of Data Sources}
The construction of our PFaithBench is based on four publicly available context-based QA datasets: HotpotQA, 2WikiMultiHopQA, SQuAD, and ConFiQA, covering diverse contextual conditions, including multi-hop reasoning, single-hop reading comprehension, and counterfactual contexts, enabling a comprehensive evaluation of pairwise faithfulness behaviors.

\textbf{HotpotQA}~\citep{yang2018hotpotqa} is a large-scale multi-hop factual QA dataset that requires models to reason over multiple supporting passages to derive the correct answer. Each question is paired with a set of Wikipedia passages, among which two or more contain the necessary supporting facts. Although HotpotQA provides sentence-level supporting fact annotations, we use passage-level contexts to preserve the difficulty of context-based reasoning. For each question, we construct the insufficient context $c^-$ by randomly replacing one of the supporting passages with an irrelevant passage. This modification breaks the multi-hop dependency while preserving the remaining contextual information.

\textbf{2WikiMultihopQA}~\citep{ho20202wiki} is another multi-hop QA dataset constructed from Wikipedia and Wikidata. Compared with HotpotQA, 2WikiMultihopQA emphasizes structured multi-hop reasoning, including comparison, inference, and compositional questions, with explicit reasoning paths derived from knowledge triples. Each question is paired with multiple supporting passages that contain the necessary evidence. We adopt the same strategy as in HotpotQA to construct insufficient contexts by randomly replacing one of the supporting passages with an irrelevant one, thereby breaking the multi-hop reasoning chain while preserving the most of contextual information.

\textbf{SQuAD}~\citep{rajpurkar2016squad} (Stanford Question Answering Dataset) is a widely used single-hop reading comprehension benchmark, in which each question is grounded in a single Wikipedia passage and the answer is a contiguous span within the passage. SQuAD primarily evaluates a model's ability to understand and extract relevant information from a given context. In our work, we use SQuAD as the source of single-hop QA data and construct insufficient contexts by rewriting the original passage to remove or blur the answer-supporting span while preserving most of the surrounding content.

\textbf{ConFiQA}~\citep{bi2025contextDPO} is a counterfactual question answering benchmark designed to evaluate the ability of LLMs to faithfully follow the provided context, even when the context contradicts real-world knowledge. Each question in ConFiQA is associated with both an original context that aligns with real-world knowledge and a counterfactual context that introduces fictitious knowledge, along with their corresponding answers. This paired design makes ConFiQA particularly suitable for studying faithfulness. In our work, we treat both the original and counterfactual contexts as sufficient contexts, and construct corresponding insufficient contexts by removing or blurring the key information through LLM-based rewriting. This results in two evaluation subsets, PFB-ConOri and PFB-ConCF, where $c^+$ corresponds to original and counterfactual contexts respectively.

\subsection{Statistics of PFaithBench}
Table~\ref{Tbl:pfb_stats} summarizes statistics of PFaithBench. We additionally construct training sets for PFB-MultiHop and PFB-SingleHop for our data-centric analyses of faithfulness training in Sec.~\ref{Sec:DataCentric_Analysis}. All evaluation samples undergo human verification. Training samples follow the same construction pipeline but are not manually verified. PFB-ConOri and PFB-ConCF are used solely for evaluation.

\subsection{Relationship between PFaithBench and Existing Benchmarks}
Table~\ref{Tbl:Position_PFaithBench} summarizes the relationship between PFaithBench and existing benchmarks from the perspectives of evaluation paradigm, context setting, and reasoning type. Existing benchmarks cover different aspects relevant to contextual faithfulness, including answerable, unanswerable, and counterfactual settings, as well as single-hop and multi-hop reasoning. However, they primarily evaluate each instance independently. In contrast, PFaithBench pairs sufficient and insufficient contexts for the same question, enabling pairwise evaluation of hether a model appropriately answers with sufficient context while abstaining with insufficient context. Meanwhile, PFaithBench covers answerable, unanswerable, and counterfactual settings across both single-hop and multi-hop reasoning, providing a unified benchmark for evaluating pairwise faithfulness under diverse contextual conditions.

\subsection{More Details of Constructing Insufficient Context $c^-$}
For multi-hop QA, HotpotQA and 2WikiMultihopQA provide annotated supporting facts. We construct $c^-$ by replacing one passage containing a key supporting fact with an irrelevant passage, thereby breaking the evidence needed to answer the question. For single-hop and counterfactual QA, we use Qwen3-32B~\citep{yang2025qwen3} to rewrite sufficient contexts $c^+$ into insufficient contexts $c^-$. Two postgraduate students then verify the evaluation examples, retaining a rewrite only if the resulting context lacks sufficient evidence to answer the question. Starting from 1,000 examples for each QA type, this process retains 762 of 1,000 single-hop examples (76.2\%) and 791 of 1,000 counterfactual examples (79.1\%). The prompt used to rewrite the contexts is shown below:
\begin{lstlisting}[title={\textbf{Prompt for rewriting insufficient contexts}}]
You will be provided with a context passage, a question, and an answer. The answer can be deduced from the given context. Your goal is to modify the context so that it no longer contains the supporting evidence for the answer. You should review the context sentence by sentence. For each sentence, consider the following two cases:
1. Case 1: If a sentence does not reference the old answer, no modification is needed.
2. Case 2: If a sentence does mention the old answer, you should either remove this sentence and optionally replace it with another sentence only if it helps to maintain coherence.

The modified context should:
1. Be coherent with the original context.
2. Not support the original answer.

Ideally, the majority of the original context should remain unchanged. You will output a JSON object containing the following information:

{{
  "question": string, // The original question.
  "old answer": string, // The original answer.
  "modified context": string, // The modified context.
  "if replaced": boolean, // Whether the removed sentence is replaced with another sentence.
  "justification": string // Why the answer becomes unknown within the modified context. The justification should be concise.
}}

Only output the JSON object. Do not output markdown fences or any extra text.

Question: {question}
Old answer: {answer}
Context passage:
{context}
\end{lstlisting}

\begin{table}[t]
\centering
\caption{\textbf{Relationship between PFaithBench and existing benchmarks.}}
\label{Tbl:Position_PFaithBench}
\vspace{-3mm}
\resizebox{0.99\textwidth}{!}{
\begin{tabular}{lccccc}
\toprule[1.5pt]
\textbf{Benchmark} &
\textbf{Evaluation Paradigm} &
\textbf{Answerable} &
\textbf{Unanswerable} &
\textbf{Counterfactual} &
\textbf{Reasoning Type} \\
\midrule
FaithEval~\citep{mingfaitheval}        & Instance-level & \checkmark & \checkmark & \checkmark & Single-hop \\
ConFiQA~\citep{bi2025contextDPO}       & Instance-level & \checkmark &             & \checkmark & Single-hop \\
TimeQA~\citep{chen2021TimeQA}           & Instance-level & \checkmark & \checkmark &             & Multi-hop \\
RGB~\citep{chen2024benchmarking}              & Instance-level & \checkmark &             &             & Single-hop \\
ClashEval~\citep{wu2024clasheval}        & Instance-level & \checkmark &             &             & Multi-hop \\
SQuAD~\citep{rajpurkar2016squad}            & Instance-level & \checkmark &             &             & Single-hop \\
NQ~\citep{kwiatkowski2019natural} & Instance-level & \checkmark &             &             & Single-hop \\
TriviaQA~\citep{joshi2017triviaqa} & Instance-level & \checkmark &             &             & Single-hop \\
PopQA~\citep{mallen2023popqa} & Instance-level & \checkmark &             &             & Single-hop \\
MuSiQue~\citep{trivedi2022musique}          & Instance-level & \checkmark &             &             & Multi-hop \\
2WikiMultihopQA~\citep{ho20202wiki}  & Instance-level & \checkmark &             &             & Multi-hop \\
HotpotQA~\citep{yang2018hotpotqa}         & Instance-level & \checkmark &             &             & Multi-hop \\
\midrule
\textbf{PFaithBench} & \textbf{Pairwise} & \checkmark & \checkmark & \checkmark &
\textbf{Both} \\
\bottomrule[1.5pt]
\end{tabular}
}
\end{table}

\begin{table}[t]
\centering
\caption{\textbf{Statistics of PFaithBench.}}
\label{Tbl:pfb_stats}
\vspace{-3mm}
\resizebox{0.99\textwidth}{!}{
\begin{tabular}{lccc}
\toprule[1.5pt]
\textbf{Subset} & \textbf{\# Eval Samples} & \textbf{\# Training Samples} & \textbf{Data Sources} \\
\midrule
PFB-MultiHop   & 1,000 & 2,000 & HotpotQA \& 2WikiMultiHopQA \\
PFB-SingleHop  & 762  & 1,997 & SQuAD \\
PFB-ConOri     & 791  & -- & ConFiQA  \\
PFB-ConCF      & 791  & -- & ConFiQA  \\
\midrule
Total          & 3,344 & 3,997 & HotpotQA, 2WikiQA, SQuAD, ConFiQA \\
\bottomrule[1.5pt]
\end{tabular}
}

\end{table}

\subsection{More Evaluation Details}
\label{App:EvaluationDetails}
We adopt a unified system prompting instruction for all models. Specifically, we explicitly instruct models to answer only based on the provided context. When the context is sufficient, the model is required to provide an answer along with its reasoning process; when the context is insufficient, the model should acknowledge the lack of information and respond with ``I don't know''. All models are evaluated with temperature 1.0 and top-$p$ 0.9. The prompt is shown below:
\begin{lstlisting}[title={\textbf{System prompt for faithfulness evaluation}}]
Answer the question based only on the provided contexts within <context> </context> tags. The contexts may or may not support to answer the question. If the contexts are sufficient, reason step by step using only the provided contexts. Put your reasoning within <think> </think> tags and your final answer within <answer> </answer> tags. If the contexts are insufficient to answer the question, do not guess. Instead, faithfully acknowledge you do not have sufficient information to answer the question within <think> </think> tags, and output "I don't know" within <answer> </answer> tags.

Output format:
<think> Your step-by-step reasoning, or a brief explanation of why the contexts are insufficient to determine the answer </think>
<answer> Your final answer or I don't know </answer>

Question: {question}
<context>
{context}
</context>
\end{lstlisting}
To compute the metrics defined in Sec.~\ref{Sec:Metrics_PairFaith}, we use Qwen3-30B-A3B-Instruct-2507~\citep{yang2025qwen3} as an LLM judge to classify model responses into $\{\mathrm{IDK},\mathrm{C},\mathrm{W}\}$. Given the question, ground-truth reference answer, and model response, the judge labels the response as abstention ($\mathrm{IDK}$), a correct answer ($\mathrm{C}$), or an incorrect answer ($\mathrm{W}$). The judge prompt is shown below:
\begin{lstlisting}[title={\textbf{Judge prompt}}]
JUDGE_INSTRUCTION = """You are an answer-comparison evaluator. Judge only whether the model response matches the provided reference answer. The reference answer is authoritative, including if it is hypothetical or counterfactual. Do not use outside knowledge, and do not judge whether the context supports the answer. Return exactly one JSON object and no other text:
{{"label":"CORRECT|INCORRECT|REFUSE","reason":"short explanation"}}

Label definitions:
- CORRECT: the response's final answer/conclusion gives the reference answer, allowing semantic paraphrases, equivalent wording, and harmless extra explanation.
- INCORRECT: the response's final answer/conclusion gives a substantive answer that does not match the reference answer.
- REFUSE: the response's final answer/conclusion provides no substantive answer and explicitly declines to answer, says it cannot determine the answer, or says "I don't know". A speculative candidate mentioned only in reasoning does not override a final refusal. If the final answer itself proposes, guesses, or states a concrete answer after a disclaimer, it is not REFUSE; compare that answer with the reference instead.

Decision procedure: determine the final answer/conclusion, giving priority to an explicit final-answer section such as <answer>...</answer> when present. Ignore speculative alternatives in the reasoning when the final conclusion is a refusal. If the final conclusion contains a concrete answer, choose CORRECT or INCORRECT by comparing it with the reference; otherwise choose REFUSE.

Question: {question}
Reference answer: {reference}
Model response: {response}
\end{lstlisting}

\subsection{Evaluation Models}
A wide range of existing LLMs is taken into the evaluation, covering seven families with thirty-nine models across parameters from 0.6 billion to 70 billion, where the most recent ones are released up to April 2026. The models include the Llama3 series (Llama-3.1/3.1/3.3-Instruct, 1B/3B/8B/70B)~\citep{grattafiori2024llama3}, Qwen3 series (0.6B/1.7B/4B/8B/14B/32B)~\citep{yang2025qwen3} and Qwen3.5/3.6 series (0.8B/2B/4B/9B/27B)~\citep{qwen3.5,qwen3.6-27b} both NoThink and Think mode, as well as Gemma series (2B/9B/27B/31B)~\citep{team2024gemma}, Phi series (3.8B/14B)~\citep{abdin2024phi3,abdin2024phi4,abouelenin2025phi4mini}, and Olmo series (7B/32B)~\citep{olmo2025olmo3}.

\begin{figure}[t]
\centering
\begin{tcolorbox}[
  colback=gray!5,
  colframe=black,
  boxrule=0.5pt,
  sharp corners,
  width=0.95\linewidth,
  title=\textbf{Example from PFaithBench}
]

\textbf{Question:} In what year did Massachusetts first require children to be educated in schools? \\

\textbf{Ground Truth Answer:} 1852 \\

\textbf{Sufficient Context:} \\
Private schooling in the United States has been debated by educators, lawmakers and parents, \textcolor{green!50!black}{since the beginnings of compulsory education in Massachusetts in 1852}. The Supreme Court precedent appears to favor educational choice, so long as states may set standards for educational accomplishment. Some of the most relevant Supreme Court case law on this is as follows: Runyon v. McCrary, 427 U.S. 160 (1976); Wisconsin v. Yoder, 406 U.S. 205 (1972); Pierce v. Society of Sisters, 268 U.S. 510 (1925); Meyer v. Nebraska, 262 U.S. 390 (1923). \\

\textbf{Insufficient Context:} \\
Private schooling in the United States has been debated by educators, lawmakers and parents, \textcolor{red}{since the beginnings of compulsory education.} The Supreme Court precedent appears to favor educational choice, so long as states may set standards for educational accomplishment. Some of the most relevant Supreme Court case law on this is as follows: Runyon v. McCrary, 427 U.S. 160 (1976); Wisconsin v. Yoder, 406 U.S. 205 (1972); Pierce v. Society of Sisters, 268 U.S. 510 (1925); Meyer v. Nebraska, 262 U.S. 390 (1923).

\end{tcolorbox}
\vspace{-2mm}
\caption{\textbf{Example of sufficient and insufficient contexts in PFaithBench.}}
\label{Fig.CaseStudy_InsufficientContext}
\end{figure}

\subsection{Case Example}
Fig.~\ref{Fig.CaseStudy_InsufficientContext} shows an example of paired sufficient and insufficient contexts in PFaithBench. The rewrite removes the key information required to answer the question while preserving the surrounding details and overall meaning of the original passage. As a result, the insufficient context remains highly similar to the sufficient context, yet no longer provides enough evidence to support the answer.

\section{Base Training Method: GRPO}
\label{App:GRPO}
The GRPO training objective under context-based question answering is as follows:
\begin{align}
\begin{split}
\label{eq:grpo}
    \mathcal{L}_\textrm{GRPO}&(\theta) = - \mathbb{E}_{(q,c,a)\in \mathcal{D}, c\in\{c^+,c^-\}, \{{y_i}\}_{i=1}^G\sim \pi_{\theta_\textrm{old}}(\cdot|x,c)}\\
    &\frac{1}{G}\sum_{i=1}^G \frac{1}{|y_i|}\sum_{t=1}^{|y_i|}\bigg( \min\bigg[ \rho_{i,t}(\theta)\hat{A}_{i,t}, \textrm{clip}(\rho_{i,t}(\theta),1-\epsilon,1+\epsilon)\hat{A}_{i,t} \bigg]-\beta\mathbb{D}_{\textrm{KL}}(\pi_\theta||\pi_\textrm{ref}) \bigg),
\end{split}
\end{align}
where
\begin{equation}
    \rho_{i,t}(\theta) = \frac{\pi_\theta(y_{i,t}|q,c,y_{i,<t})}{\pi_{\theta_\textrm{old}}(y_{i,t}|q,c,y_{i,<t})},\; \hat{A}_{i,t} = \frac{r(a,y_i) - \text{mean}(\{r(a,y_i)\}_{i=1}^{G})}{\text{std}(\{r(a,y_i)\}_{i=1}^{G})}.
\end{equation}
Here, $\rho_{i,t}(\theta)$ denotes the policy ratio of importance sampling, $\hat{A}_{i,t}$ is the normalized advantage, and $\mathbb{D}_{\textrm{KL}}$ is the KL divergence term. We use it as the training algorithm in our data-centric analysis.

\begin{table}[t]
\centering
\caption{\textbf{Imbalance and asymmetry on PFaithBench across diverse LLMs.} Imbalance is $|\mathrm{AnsAcc}-\mathrm{AbsAcc}|$; asymmetry is $|\max(\mathrm{AnsAcc},\mathrm{AbsAcc})-\mathrm{SFAcc}|$. Avg. is the mean across four subsets. The minimum in each model family and column is highlighted in \textbf{bold}.}
\label{Tble:Imbalance_Asymmetry}
\vspace{-3mm}
\resizebox{\textwidth}{!}{
\begin{threeparttable}
\begin{tabular}{ll|cc|cc|cc|cc|cc}
    \toprule[1.5pt]
    \multirow{2}*{\textbf{Family}} & \multirow{2}*{\textbf{LLM}} & \multicolumn{2}{c}{\textbf{PFB-MultiHop}} & \multicolumn{2}{c}{\textbf{PFB-SingleHop}} & \multicolumn{2}{c}{\textbf{PFB-ConOri}} & \multicolumn{2}{c}{\textbf{PFB-ConCF}} & \multicolumn{2}{c}{\textbf{Avg.}} \\
    \cmidrule(lr){3-4} \cmidrule(lr){5-6} \cmidrule(lr){7-8} \cmidrule(lr){9-10} \cmidrule(lr){11-12}
    ~ & ~ & \textbf{Imbalance} $\downarrow$ & \textbf{Asymmetry} $\downarrow$ & \textbf{Imbalance} $\downarrow$ & \textbf{Asymmetry} $\downarrow$ & \textbf{Imbalance} $\downarrow$ & \textbf{Asymmetry} $\downarrow$ & \textbf{Imbalance} $\downarrow$ & \textbf{Asymmetry} $\downarrow$ & \textbf{Imbalance} $\downarrow$ & \textbf{Asymmetry} $\downarrow$ \\
    \midrule
    \multirow{5}*{\textbf{Llama3}} & Llama-3.2-1B-Instruct & 27.50 & 40.50 & 25.33 & 40.55 & 34.26 & 48.55 & 49.43 & 51.83 & 34.13 & 45.36 \\
    ~ & Llama-3.2-3B-Instruct & \textbf{16.70} & 36.60 & 54.72 & 61.15 & 22.88 & 36.66 & 60.68 & 62.96 & 38.75 & 49.34 \\
    ~ & Llama-3.1-8B-Instruct & 30.40 & 43.80 & 18.77 & 33.86 & 30.34 & 39.32 & \textbf{23.39} & \textbf{33.88} & 25.72 & 37.71 \\
    ~ & Llama-3.1-70B-Instruct & 19.30 & \textbf{32.10} & \textbf{7.87} & \textbf{17.98} & \textbf{13.65} & \textbf{26.42} & 39.19 & 47.03 & \textbf{20.00} & \textbf{30.88} \\
    ~ & Llama-3.3-70B-Instruct & 18.80 & 32.30 & 16.40 & 24.02 & 18.58 & 28.57 & 32.36 & 40.58 & 21.54 & 31.37 \\
    \midrule
    \multirow{6}*{\textbf{Qwen3 (NoThink)}} & Qwen3-0.6B (NoThink) & 6.60 & 30.70 & 9.71 & 31.50 & 24.53 & 37.17 & \textbf{14.54} & 32.62 & 13.84 & 33.00 \\
    ~ & Qwen3-1.7B (NoThink) & \textbf{4.80} & 31.60 & 8.27 & 31.89 & 19.72 & 34.64 & 16.43 & \textbf{31.86} & \textbf{12.31} & 32.50 \\
    ~ & Qwen3-4B (NoThink) & 10.30 & 34.20 & 6.56 & 23.10 & 13.78 & 26.30 & 35.15 & 42.48 & 16.45 & 31.52 \\
    ~ & Qwen3-8B (NoThink) & 21.30 & 38.90 & \textbf{0.13} & \textbf{18.64} & \textbf{2.53} & \textbf{16.06} & 45.89 & 50.44 & 17.46 & 31.01 \\
    ~ & Qwen3-14B (NoThink) & 7.00 & \textbf{28.40} & 11.29 & 19.95 & 10.49 & 19.60 & 23.77 & 34.01 & 13.14 & \textbf{25.49} \\
    ~ & Qwen3-32B (NoThink) & 23.20 & 35.20 & 27.95 & 32.41 & 17.83 & 26.30 & 20.86 & 33.00 & 22.46 & 31.73 \\
    \midrule
    \multirow{6}*{\textbf{Qwen3 (Think)}} & Qwen3-0.6B (Think) & 46.80 & 53.50 & 66.93 & 69.29 & 69.03 & 71.81 & 31.86 & 38.56 & 53.65 & 58.29 \\
    ~ & Qwen3-1.7B (Think) & 48.60 & 55.60 & 67.45 & 69.16 & 69.91 & 72.95 & 30.85 & 37.80 & 54.20 & 58.88 \\
    ~ & Qwen3-4B (Think) & \textbf{27.50} & \textbf{39.20} & 30.45 & 36.75 & \textbf{28.07} & 37.29 & 8.60 & 26.30 & 23.65 & 34.88 \\
    ~ & Qwen3-8B (Think) & 27.70 & \textbf{39.20} & \textbf{24.02} & \textbf{30.84} & 28.19 & \textbf{36.54} & 9.73 & 26.55 & \textbf{22.41} & \textbf{33.28} \\
    ~ & Qwen3-14B (Think) & 28.70 & 40.10 & 35.83 & 40.94 & 35.90 & 43.99 & \textbf{6.07} & \textbf{25.16} & 26.62 & 37.55 \\
    ~ & Qwen3-32B (Think) & 37.90 & 46.40 & 41.86 & 45.14 & 50.70 & 56.51 & 13.53 & 30.09 & 36.00 & 44.54 \\
    \midrule
    \multirow{5}*{\textbf{Qwen3.5 (NoThink)}} & Qwen3.5-0.8B (NoThink) & \textbf{13.10} & \textbf{36.90} & \textbf{9.32} & 33.33 & \textbf{1.77} & \textbf{19.97} & 36.92 & 44.88 & \textbf{15.28} & 33.77 \\
    ~ & Qwen3.5-2B (NoThink) & 31.70 & 45.10 & 25.33 & 39.24 & 12.64 & 25.41 & 22.12 & \textbf{33.63} & 22.95 & 35.84 \\
    ~ & Qwen3.5-4B (NoThink) & 33.50 & 45.20 & 20.08 & 28.35 & 18.58 & 26.68 & 24.78 & 38.18 & 24.24 & 34.60 \\
    ~ & Qwen3.5-9B (NoThink) & 29.90 & 40.30 & 24.80 & 29.40 & 13.91 & 24.53 & \textbf{22.00} & 35.78 & 22.65 & 32.50 \\
    ~ & Qwen3.5-27B (NoThink) & 32.90 & 43.30 & 22.05 & \textbf{25.59} & 11.38 & 20.73 & 29.84 & 40.08 & 24.04 & \textbf{32.42} \\
    \midrule
    \multirow{5}*{\textbf{Qwen3.5 (Think)}} & Qwen3.5-0.8B (Think) & 41.50 & 51.70 & 40.68 & 50.39 & 36.16 & 46.65 & 6.70 & 24.40 & 31.26 & 43.29 \\
    ~ & Qwen3.5-2B (Think) & 43.90 & 52.30 & 26.12 & 33.86 & 18.84 & 29.08 & 20.35 & 34.13 & 27.30 & 37.34 \\
    ~ & Qwen3.5-4B (Think) & 58.10 & 65.30 & 48.16 & 52.10 & 37.93 & 45.01 & \textbf{0.63} & \textbf{22.88} & 36.21 & 46.32 \\
    ~ & Qwen3.5-9B (Think) & \textbf{16.20} & \textbf{28.70} & \textbf{8.40} & \textbf{16.14} & \textbf{4.42} & \textbf{11.63} & 37.17 & 41.34 & \textbf{16.55} & \textbf{24.45} \\
    ~ & Qwen3.5-27B (Think) & 64.30 & 71.20 & 42.91 & 46.46 & 26.68 & 35.40 & 9.48 & 28.19 & 35.84 & 45.31 \\
    \midrule
    \multirow{2}*{\textbf{Qwen3.6}} & Qwen3.6-27B (NoThink) & \textbf{25.20} & \textbf{36.90} & \textbf{20.60} & \textbf{24.93} & \textbf{10.75} & \textbf{20.99} & 31.10 & 41.59 & \textbf{21.91} & \textbf{31.10} \\
    ~ & Qwen3.6-27B (Think) & 65.20 & 69.50 & 50.26 & 51.71 & 34.89 & 41.21 & \textbf{1.64} & \textbf{28.32} & 38.00 & 47.68 \\
    \midrule
    \multirow{4}*{\textbf{Gemma}} & Gemma-2-2B-It & 37.10 & 48.40 & \textbf{11.81} & 30.05 & 7.84 & 26.80 & 29.08 & 39.70 & 21.46 & 36.24 \\
    ~ & Gemma-2-9B-It & 45.90 & 56.00 & 31.10 & 37.66 & 35.02 & 42.10 & \textbf{5.06} & \textbf{24.91} & 29.27 & 40.17 \\
    ~ & Gemma-2-27B-It & 29.60 & 44.30 & 18.64 & 25.46 & 12.26 & 21.62 & 26.93 & 35.90 & 21.86 & 31.82 \\
    ~ & Gemma-4-31B-it & \textbf{23.30} & \textbf{34.90} & 14.04 & \textbf{18.77} & \textbf{1.14} & \textbf{14.79} & 40.33 & 47.66 & \textbf{19.70} & \textbf{29.03} \\
    \midrule
    \multirow{4}*{\textbf{Phi}} & Phi-3-mini-128k-instruct & 16.20 & 39.20 & 27.82 & 41.08 & \textbf{5.06} & \textbf{20.61} & 53.86 & 56.51 & 25.73 & 39.35 \\
    ~ & Phi-3.5-mini-instruct & 11.40 & 36.80 & 13.78 & 31.10 & 11.13 & 25.16 & \textbf{37.80} & \textbf{43.49} & 18.53 & 34.14 \\
    ~ & Phi-4-mini-instruct & \textbf{4.20} & \textbf{28.60} & \textbf{7.87} & 25.85 & 8.60 & 20.86 & 39.06 & 46.65 & \textbf{14.93} & \textbf{30.49} \\
    ~ & Phi-4 & 22.00 & 36.20 & 16.01 & \textbf{25.72} & 7.96 & 22.63 & 45.51 & 48.93 & 22.87 & 33.37 \\
    \midrule
    \multirow{2}*{\textbf{Olmo}} & Olmo-3-7B-Instruct & 49.60 & 58.00 & 55.91 & 59.58 & 30.34 & 37.80 & 63.84 & 66.12 & 49.92 & 55.37 \\
    ~ & Olmo-3.1-32B-Instruct & \textbf{11.20} & \textbf{28.90} & \textbf{7.09} & \textbf{17.59} & \textbf{3.29} & \textbf{16.18} & \textbf{40.96} & \textbf{46.78} & \textbf{15.63} & \textbf{27.36} \\
    \bottomrule[1.5pt]
\end{tabular}
\end{threeparttable}}
\end{table}

\section{Further Experimental Results}
\subsection{More Metrics of Imbalance and Asymmetry}
\textbf{Imbalance and asymmetry.} Under our pairwise setting, $c^+$ and $c^-$ differ only in the key information required to answer the question, so a faithful model should maintain consistent answering and abstaining capabilities across the paired contexts. We introduce two diagnostic metrics: \emph{Imbalance} to measure the discrepancy between marginal answering and abstention performance, and \emph{Asymmetry} to measure the gap between the stronger marginal performance and strict pairwise faithfulness:
\begin{equation}
    \mathrm{Imbalance} = |\mathrm{AnsAcc}-\mathrm{AbsAcc}|, \quad
    \mathrm{Asymmetry} = |\max(\mathrm{AnsAcc},\mathrm{AbsAcc})-\mathrm{SFAcc}|.
\end{equation}
Higher values indicate more severe imbalance between answering and abstaining capabilities or a larger discrepancy between marginal capability and pairwise faithfulness, respectively.

\par
Table~\ref{Tble:Imbalance_Asymmetry} reports the results of the two metrics. We observe substantial imbalance and asymmetry across diverse model families and scales, indicating that strong marginal answering or abstaining capability does not necessarily translate into consistent behavior over paired contexts. In particular, asymmetry remains pronounced even when imbalance is relatively small, suggesting that models may exhibit seemingly balanced marginal capabilities while still failing to answer under sufficient context and abstain under insufficient context for the same question. Moreover, increasing model scale does not consistently reduce either metric, showing that such discrepancies are not naturally resolved by scaling. A similar pattern holds for reasoning-enhanced models: enabling thinking does not reliably improve imbalance or asymmetry and, in many cases, further enlarges them. These results further highlight the limitation of evaluating answering and abstaining capabilities independently and motivate pairwise evaluation for revealing faithfulness failures that marginal metrics can obscure.

\begin{figure}[t]
  \centering
  \vspace{2mm}
  \includegraphics[width=0.325\textwidth]{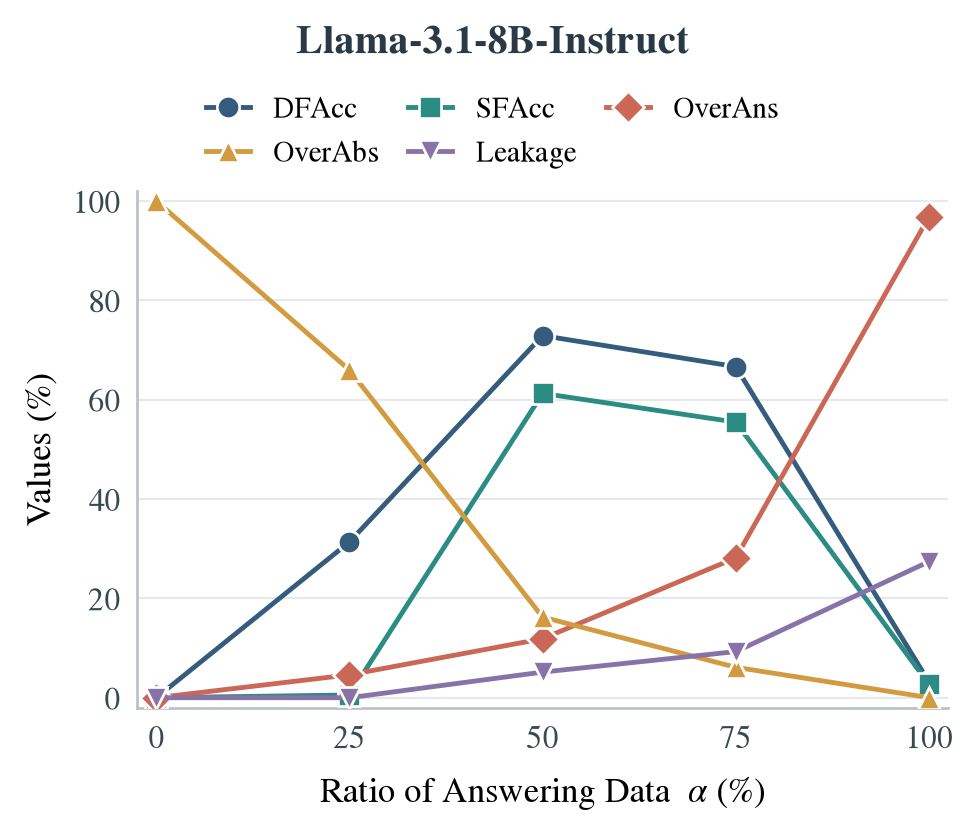}
  \includegraphics[width=0.325\textwidth]{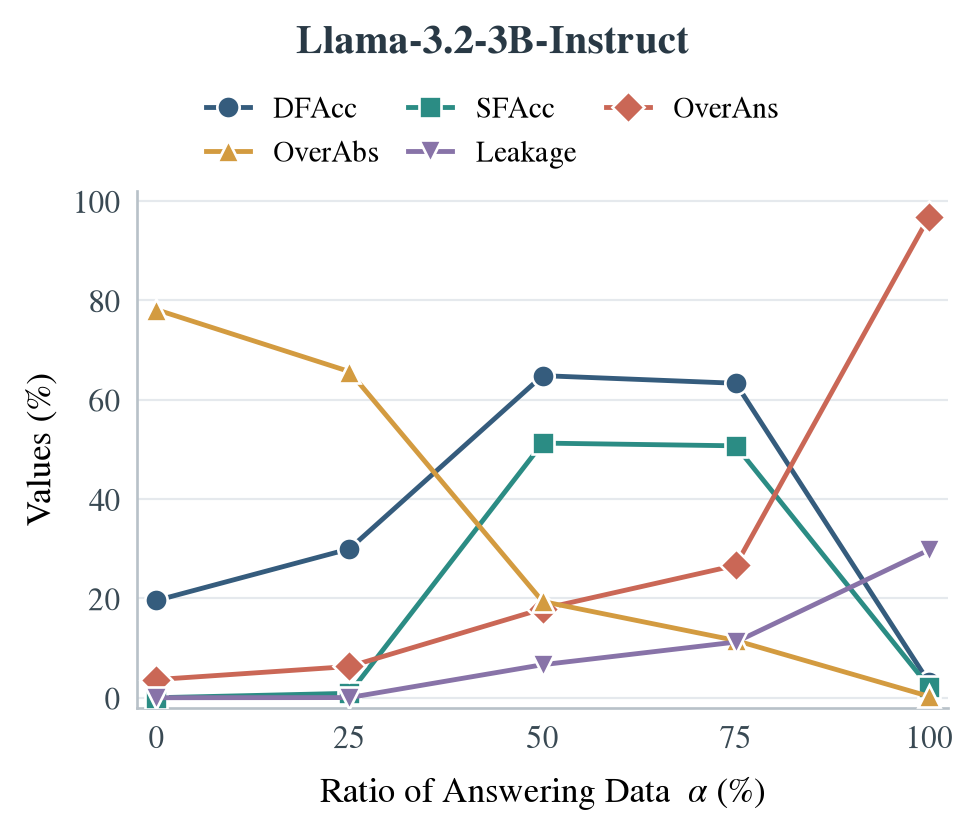}
  \includegraphics[width=0.325\textwidth]{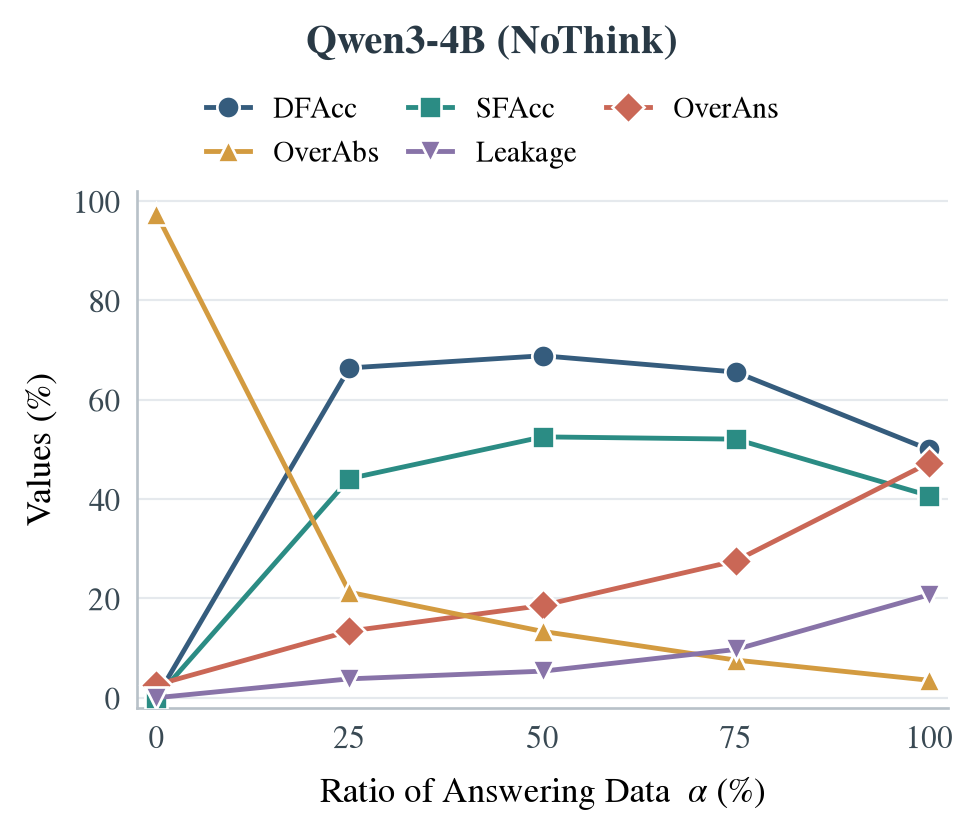}
  \vspace{-3mm}
  \caption{\textbf{Impact of answering–abstention data composition (symmetric reward).} Faithfulness training under different answering data ratios $\alpha \in \{0,25,50,75,100\}$ across three LLMs. This is the average value across four subsets in PFaithBench. 
  }
  \label{Fig:data_composition_symmetric}
\end{figure}

\begin{table}[t!]
\centering
\caption{\textbf{Family-level performance across base models on PFaithBench.} Metrics are averaged equally across models and the four subsets. Average model sizes are in billions of parameters.}
\label{Tble:ModelFamilySummary}
\vspace{-3mm}
\resizebox{\textwidth}{!}{
\begin{threeparttable}
\begin{tabular}{l|c|cccccc}
\toprule[1.5pt]
\textbf{Model Family} & \textbf{Avg. Model Size (B)} & \textbf{SFAcc} $\uparrow$ & \textbf{OverAns} $\downarrow$ & \textbf{OverAbs} $\downarrow$ & \textbf{AnsAcc} $\uparrow$ & \textbf{AbsAcc} $\uparrow$ & \textbf{Leakage} $\downarrow$ \\
\midrule
Llama & 30.40 Billion & 33.36 & 36.41 & 19.48 & 52.97 & 63.59 & 14.67 \\
Qwen3 (NoThink) & 10.05 Billion & 42.64 & 33.73 & 14.52 & 64.82 & 66.27 & 10.34 \\
Qwen3 (Think) & 10.05 Billion & 31.09 & 59.66 & 3.67 & 74.90 & 40.34 & 14.66 \\
Qwen3.5 (NoThink) & 8.56 Billion & 45.47 & 35.28 & 7.06 & 72.05 & 64.72 & 8.80 \\
Qwen3.5 (Think) & 8.56 Billion & 39.58 & 46.57 & 5.06 & 74.98 & 53.43 & 6.92 \\
Qwen3.6 & 27.00 Billion & 44.47 & 42.21 & 4.26 & 79.98 & 57.79 & 12.67 \\
Gemma & 17.25 Billion & 42.91 & 36.45 & 8.16 & 67.83 & 63.55 & 8.81 \\
Phi & 6.35 Billion & 41.75 & 28.99 & 15.09 & 60.65 & 71.01 & 11.80 \\
Olmo & 19.50 Billion & 44.06 & 14.57 & 33.21 & 52.65 & 85.43 & 5.81 \\
\bottomrule[1.5pt]
\end{tabular}
\end{threeparttable}}
\end{table}

\subsection{More Results of Symmetric Reward}
\label{App:Symmetric_reward}
As a supplement to Sec.~\ref{Sec:Asymmetric_reward}, where we adopt an asymmetric reward that penalizes abstention more strongly than incorrect answering under sufficient context, we further investigate a symmetric reward scheme that assigns equal penalties to incorrect answers and abstention. Specifically, the reward is:
\begin{equation}
r(a,y^-)=
\begin{cases}
+1 & \text{if } y^-=\mathrm{IDK},\\
-1 & \text{if } y^-\in\{\mathrm{C},\mathrm{W}\},
\end{cases}
\quad
r_{\mathrm{sym}}(a,y^+)=
\begin{cases}
+1 & \text{if } y^+=\mathrm{C},\\
-1 & \text{if } y^+\in\{\mathrm{W},\mathrm{IDK}\},
\end{cases}
\end{equation}
while keeping the reward under insufficient context $c^-$ unchanged. Fig.~\ref{Fig:data_composition_symmetric} illustrates the impact of the answering-abstention data composition under the symmetric reward. Overall, we observe a trend qualitatively consistent with that under the asymmetric reward in Fig.~\ref{Fig:data_composition_asymmetric}. When $\alpha=0$, training exclusively on abstention data leads to severe over-abstention and poor pairwise faithfulness. As $\alpha$ increases, over-abstention decreases substantially, while DFAcc and SFAcc improve and generally reach their best performance with a moderate mixture of answering and abstention data. In contrast, when the training data becomes dominated by answering examples, particularly at $\alpha=100$, over-answering increases and pairwise faithfulness deteriorates. These results confirm that balancing answering and abstention supervision is important for faithfulness training, and that the data-composition effect observed in the main experiments is robust to the choice of reward design.

\subsection{Family-level Results}
Table~\ref{Tble:ModelFamilySummary} summarizes the performance of different model families on PFaithBench. Most model families exhibit a clear tendency toward over-answering, with OverAns noticeably higher than OverAbs; Olmo is the main exception, showing a stronger tendency toward over-abstention. This indicates that existing LLMs generally tend to answer even when the provided context is insufficient, revealing over-answering as a prevalent failure mode of pairwise faithfulness. Moreover, pairwise faithfulness shows no clear positive relationship with average model size. Smaller families such as Qwen3.5 and Phi achieve comparable or even higher SFAcc than substantially larger families, suggesting that increasing model scale alone does not naturally resolve pairwise faithfulness failures.

\begin{table}[t]
\centering
\caption{\textbf{Detailed impact of answering-abstention data composition (asymmetric reward).} The answering data percentage $\alpha$ vary $\alpha\in\{0,25,50,75,100\}$, serving as a supplement to Fig.~\ref{Fig:data_composition_asymmetric}.}
\label{Tble:Datacomposition_Supplement}
\vspace{-3mm}
\resizebox{\textwidth}{!}{
\begin{threeparttable}
\begin{tabular}{ll|ccccc|ccccc}
    \toprule[1.5pt]
    \multirow{2}*{\textbf{LLM}} & \multirow{2}*{$\alpha$} & \multicolumn{5}{c}{\textbf{PFB-MultiHop}} & \multicolumn{5}{c}{\textbf{PFB-SingleHop}} \\
    \cmidrule(lr){3-7} \cmidrule(lr){8-12}
    ~ & ~ & \textbf{DFAcc} $\uparrow$ & \textbf{SFAcc} $\uparrow$ & \textbf{OverAns} $\downarrow$ & \textbf{OverAbs} $\downarrow$ & \textbf{Leakage} $\downarrow$ & \textbf{DFAcc} $\uparrow$ & \textbf{SFAcc} $\uparrow$ & \textbf{OverAns} $\downarrow$ & \textbf{OverAbs} $\downarrow$ & \textbf{Leakage} $\downarrow$ \\
    \midrule
    \multirow{5}*{\textbf{Llama-3.1-8B-Instruct}} & $\alpha=0$ & 10.90 & 0.00 & 10.50 & 84.70 & 0.00 & 18.50 & 0.00 & 2.23 & 80.05 & 0.00 \\
    ~ & $\alpha=25$ & 60.60 & 47.90 & 28.80 & 12.10 & 9.10 & 76.25 & 65.09 & 9.58 & 14.96 & 0.79 \\
    ~ & $\alpha=50$ & 69.10 & 59.10 & 17.80 & 13.60 & 10.40 & 76.51 & 73.36 & 13.25 & 10.50 & 1.44 \\
    ~ & $\alpha=75$ & 58.50 & 49.40 & 36.10 & 5.90 & 18.50 & 62.99 & 59.71 & 34.51 & 3.02 & 2.89 \\
    ~ & $\alpha=100$ & 0.40 & 0.40 & 99.60 & 0.00 & 52.80 & 0.26 & 0.26 & 99.74 & 0.00 & 12.73 \\
    \midrule
    \multirow{5}*{\textbf{Llama-3.2-3B-Instruct}} & $\alpha=0$ & 15.20 & 0.00 & 10.30 & 79.50 & 0.00 & 12.86 & 0.00 & 0.66 & 86.88 & 0.00 \\
    ~ & $\alpha=25$ & 53.70 & 35.20 & 19.60 & 28.60 & 6.80 & 54.33 & 45.41 & 6.30 & 40.55 & 0.92 \\
    ~ & $\alpha=50$ & 59.60 & 45.60 & 26.70 & 15.40 & 13.10 & 63.25 & 56.04 & 21.00 & 17.06 & 1.71 \\
    ~ & $\alpha=75$ & 35.80 & 28.50 & 61.60 & 3.20 & 29.30 & 37.27 & 34.91 & 60.50 & 2.23 & 4.99 \\
    ~ & $\alpha=100$ & 0.00 & 0.00 & 100.00 & 0.00 & 48.10 & 0.26 & 0.26 & 99.74 & 0.00 & 10.89 \\
    \midrule
    \multirow{5}*{\textbf{Qwen3-4B (NoThink)}} & $\alpha=0$ & 0.00 & 0.00 & 3.20 & 96.80 & 0.00 & 0.00 & 0.00 & 0.00 & 100.00 & 0.00 \\
    ~ & $\alpha=25$ & 57.20 & 43.10 & 31.20 & 12.90 & 9.60 & 71.13 & 55.38 & 7.74 & 21.52 & 0.13 \\
    ~ & $\alpha=50$ & 57.20 & 46.40 & 35.20 & 8.40 & 14.80 & 71.13 & 63.39 & 19.03 & 10.37 & 1.44 \\
    ~ & $\alpha=75$ & 51.00 & 42.60 & 44.20 & 5.30 & 20.50 & 62.73 & 57.35 & 33.86 & 3.81 & 6.30 \\
    ~ & $\alpha=100$ & 25.30 & 21.40 & 74.20 & 0.80 & 40.60 & 25.33 & 23.88 & 73.88 & 0.79 & 13.25 \\
    \midrule
    \multirow{2}*{\textbf{LLM}} & \multirow{2}*{$\alpha$} & \multicolumn{5}{c}{\textbf{PFB-ConOri}} & \multicolumn{5}{c}{\textbf{PFB-ConCF}} \\
    \cmidrule(lr){3-7} \cmidrule(lr){8-12}
    ~ & ~ & \textbf{DFAcc} $\uparrow$ & \textbf{SFAcc} $\uparrow$ & \textbf{OverAns} $\downarrow$ & \textbf{OverAbs} $\downarrow$ & \textbf{Leakage} $\downarrow$ & \textbf{DFAcc} $\uparrow$ & \textbf{SFAcc} $\uparrow$ & \textbf{OverAns} $\downarrow$ & \textbf{OverAbs} $\downarrow$ & \textbf{Leakage} $\downarrow$ \\
    \midrule
    \multirow{5}*{\textbf{Llama-3.1-8B-Instruct}} & $\alpha=0$ & 36.41 & 0.00 & 2.40 & 61.69 & 0.00 & 25.54 & 0.00 & 5.18 & 71.05 & 0.00 \\
    ~ & $\alpha=25$ & 76.11 & 67.00 & 13.27 & 11.38 & 2.53 & 64.35 & 31.10 & 13.65 & 24.27 & 0.00 \\
    ~ & $\alpha=50$ & 81.16 & 76.99 & 8.60 & 10.62 & 1.52 & 66.75 & 39.44 & 12.64 & 23.26 & 0.13 \\
    ~ & $\alpha=75$ & 60.43 & 55.37 & 35.40 & 4.55 & 9.99 & 49.05 & 30.72 & 43.49 & 9.86 & 0.25 \\
    ~ & $\alpha=100$ & 0.38 & 0.38 & 99.62 & 0.00 & 29.96 & 0.38 & 0.00 & 99.62 & 0.00 & 1.77 \\
    \midrule
    \multirow{5}*{\textbf{Llama-3.2-3B-Instruct}} & $\alpha=0$ & 23.77 & 0.00 & 2.78 & 74.59 & 0.00 & 25.41 & 0.00 & 3.03 & 71.93 & 0.00 \\
    ~ & $\alpha=25$ & 73.32 & 65.11 & 9.61 & 17.19 & 5.56 & 45.26 & 24.15 & 9.73 & 48.80 & 0.13 \\
    ~ & $\alpha=50$ & 75.09 & 69.41 & 15.42 & 9.86 & 9.36 & 57.90 & 34.89 & 16.06 & 30.85 & 0.25 \\
    ~ & $\alpha=75$ & 53.86 & 51.07 & 42.60 & 3.67 & 19.22 & 51.33 & 30.97 & 42.10 & 7.96 & 0.51 \\
    ~ & $\alpha=100$ & 0.25 & 0.25 & 99.75 & 0.00 & 32.87 & 1.14 & 0.38 & 98.86 & 0.00 & 0.88 \\
    \midrule
    \multirow{5}*{\textbf{Qwen3-4B (NoThink)}} & $\alpha=0$ & 0.00 & 0.00 & 2.53 & 97.47 & 0.00 & 0.00 & 0.00 & 4.30 & 95.70 & 0.00 \\
    ~ & $\alpha=25$ & 75.98 & 66.12 & 11.13 & 13.15 & 2.40 & 62.58 & 26.55 & 10.75 & 28.07 & 0.00 \\
    ~ & $\alpha=50$ & 75.73 & 66.88 & 15.93 & 8.72 & 5.56 & 65.87 & 35.02 & 17.07 & 19.97 & 0.00 \\
    ~ & $\alpha=75$ & 66.37 & 61.95 & 28.82 & 5.18 & 13.15 & 60.68 & 35.15 & 28.45 & 14.54 & 0.51 \\
    ~ & $\alpha=100$ & 36.54 & 33.63 & 62.07 & 1.39 & 29.46 & 33.38 & 19.97 & 63.08 & 5.06 & 0.76 \\
    \bottomrule[1.5pt]
\end{tabular}
\end{threeparttable}}
\end{table}

\begin{table}[t]
\centering
\caption{\textbf{Detailed impact of answering-abstention data composition (symmetric reward).} The answering data percentage $\alpha$ vary $\alpha\in\{0,25,50,75,100\}$, serving as a supplement to Fig.~\ref{Fig:data_composition_symmetric}.}
\label{Tble:Datacomposition_Symmetric_Supplement}
\vspace{-3mm}
\resizebox{\textwidth}{!}{
\begin{threeparttable}
\begin{tabular}{ll|ccccc|ccccc}
    \toprule[1.5pt]
    \multirow{2}*{\textbf{LLM}} & \multirow{2}*{$\alpha$} & \multicolumn{5}{c}{\textbf{PFB-MultiHop}} & \multicolumn{5}{c}{\textbf{PFB-SingleHop}} \\
    \cmidrule(lr){3-7} \cmidrule(lr){8-12}
    ~ & ~ & DFAcc $\uparrow$ & SFAcc $\uparrow$ & OverAns $\downarrow$ & OverAbs $\downarrow$ & Leakage $\downarrow$ & DFAcc $\uparrow$ & SFAcc $\uparrow$ & OverAns $\downarrow$ & OverAbs $\downarrow$ & Leakage $\downarrow$ \\
    \midrule
    \multirow{5}*{\textbf{Llama-3.1-8B-Instruct}} & $\alpha=0$ & 0.00 & 0.00 & 0.00 & 100.00 & 0.00 & 0.00 & 0.00 & 0.00 & 100.00 & 0.00 \\
    ~ & $\alpha=25$ & 17.00 & 0.30 & 9.30 & 78.90 & 0.00 & 24.67 & 0.52 & 1.97 & 74.67 & 0.00 \\
    ~ & $\alpha=50$ & 68.10 & 58.70 & 17.10 & 15.40 & 11.90 & 78.35 & 72.97 & 11.68 & 10.24 & 1.44 \\
    ~ & $\alpha=75$ & 63.90 & 53.60 & 30.00 & 6.30 & 18.90 & 71.39 & 67.59 & 26.38 & 2.76 & 3.02 \\
    ~ & $\alpha=100$ & 2.70 & 2.10 & 97.30 & 0.10 & 51.50 & 3.67 & 3.54 & 96.33 & 0.00 & 9.32 \\
    \midrule
    \multirow{5}*{\textbf{Llama-3.2-3B-Instruct}} & $\alpha=0$ & 14.50 & 0.00 & 8.90 & 80.40 & 0.00 & 12.73 & 0.00 & 0.66 & 86.88 & 0.00 \\
    ~ & $\alpha=25$ & 25.50 & 0.90 & 14.70 & 64.90 & 0.10 & 22.05 & 0.26 & 2.76 & 76.12 & 0.00 \\
    ~ & $\alpha=50$ & 58.80 & 43.30 & 26.80 & 16.60 & 12.20 & 67.45 & 56.82 & 14.30 & 19.69 & 1.18 \\
    ~ & $\alpha=75$ & 54.30 & 42.00 & 36.80 & 10.00 & 19.10 & 63.91 & 56.82 & 28.61 & 8.79 & 3.02 \\
    ~ & $\alpha=100$ & 0.70 & 0.30 & 99.30 & 0.10 & 46.80 & 1.44 & 1.31 & 98.56 & 0.13 & 9.97 \\
    \midrule
    \multirow{5}*{\textbf{Qwen3-4B (NoThink)}} & $\alpha=0$ & 0.00 & 0.00 & 3.50 & 96.50 & 0.00 & 0.00 & 0.00 & 0.00 & 100.00 & 0.00 \\
    ~ & $\alpha=25$ & 58.10 & 43.70 & 28.10 & 15.50 & 9.50 & 66.01 & 45.41 & 7.22 & 27.56 & 0.39 \\
    ~ & $\alpha=50$ & 57.20 & 45.20 & 33.90 & 9.60 & 12.10 & 75.72 & 65.62 & 12.47 & 12.60 & 0.66 \\
    ~ & $\alpha=75$ & 50.10 & 41.30 & 44.60 & 5.60 & 18.40 & 69.55 & 63.52 & 25.59 & 5.38 & 3.15 \\
    ~ & $\alpha=100$ & 39.20 & 33.90 & 58.40 & 2.80 & 32.50 & 49.48 & 45.41 & 49.61 & 1.18 & 8.66 \\
    \midrule
    \multirow{2}*{\textbf{LLM}} & \multirow{2}*{$\alpha$} & \multicolumn{5}{c}{\textbf{PFB-ConOri}} & \multicolumn{5}{c}{\textbf{PFB-ConCF}} \\
    \cmidrule(lr){3-7} \cmidrule(lr){8-12}
    ~ & ~ & DFAcc $\uparrow$ & SFAcc $\uparrow$ & OverAns $\downarrow$ & OverAbs $\downarrow$ & Leakage $\downarrow$ & DFAcc $\uparrow$ & SFAcc $\uparrow$ & OverAns $\downarrow$ & OverAbs $\downarrow$ & Leakage $\downarrow$ \\
    \midrule
    \multirow{5}*{\textbf{Llama-3.1-8B-Instruct}} & $\alpha=0$ & 0.00 & 0.00 & 0.00 & 100.00 & 0.00 & 0.00 & 0.00 & 0.00 & 100.00 & 0.00 \\
    ~ & $\alpha=25$ & 45.51 & 1.39 & 2.65 & 52.34 & 0.00 & 38.31 & 0.00 & 4.55 & 57.90 & 0.00 \\
    ~ & $\alpha=50$ & 80.66 & 75.85 & 8.72 & 10.87 & 2.15 & 64.48 & 37.55 & 9.99 & 28.45 & 0.25 \\
    ~ & $\alpha=75$ & 70.29 & 64.98 & 26.17 & 3.67 & 5.94 & 61.06 & 35.78 & 30.09 & 11.76 & 0.00 \\
    ~ & $\alpha=100$ & 3.41 & 3.16 & 96.59 & 0.00 & 21.37 & 3.03 & 2.15 & 96.97 & 0.00 & 1.14 \\
    \midrule
    \multirow{5}*{\textbf{Llama-3.2-3B-Instruct}} & $\alpha=0$ & 24.27 & 0.00 & 2.15 & 74.34 & 0.00 & 26.80 & 0.00 & 2.91 & 70.92 & 0.00 \\
    ~ & $\alpha=25$ & 38.05 & 2.28 & 3.29 & 59.42 & 0.00 & 34.26 & 0.13 & 4.42 & 62.20 & 0.00 \\
    ~ & $\alpha=50$ & 76.61 & 71.05 & 13.53 & 10.24 & 6.57 & 56.51 & 33.88 & 16.94 & 30.97 & 0.25 \\
    ~ & $\alpha=75$ & 73.07 & 67.76 & 20.35 & 6.83 & 11.50 & 61.95 & 36.28 & 20.86 & 20.48 & 0.25 \\
    ~ & $\alpha=100$ & 5.44 & 4.80 & 94.44 & 0.38 & 32.49 & 4.93 & 2.53 & 94.82 & 0.51 & 1.14 \\
    \midrule
    \multirow{5}*{\textbf{Qwen3-4B (NoThink)}} & $\alpha=0$ & 0.00 & 0.00 & 2.53 & 97.60 & 0.00 & 0.00 & 0.00 & 4.30 & 95.70 & 0.00 \\
    ~ & $\alpha=25$ & 76.99 & 61.44 & 9.48 & 13.78 & 1.52 & 64.48 & 25.92 & 8.85 & 28.07 & 0.00 \\
    ~ & $\alpha=50$ & 75.85 & 65.36 & 13.65 & 10.49 & 3.29 & 66.62 & 33.88 & 14.29 & 20.73 & 0.00 \\
    ~ & $\alpha=75$ & 75.47 & 68.27 & 18.58 & 6.07 & 7.59 & 67.13 & 35.15 & 21.49 & 13.27 & 0.00 \\
    ~ & $\alpha=100$ & 57.77 & 53.22 & 39.57 & 2.78 & 20.99 & 53.48 & 30.21 & 41.47 & 7.33 & 0.51 \\
    \bottomrule[1.5pt]
\end{tabular}
\end{threeparttable}}
\end{table}

\subsection{More Results of Answering-abstention Data Composition.}
As a supplement to Fig.~\ref{Fig:data_composition_asymmetric} and Fig.~\ref{Fig:data_composition_symmetric}, Tables~\ref{Tble:Datacomposition_Supplement} and~\ref{Tble:Datacomposition_Symmetric_Supplement} provide the detailed results across all four PFaithBench subsets under the asymmetric and symmetric rewards, respectively. The results consistently show that using only answering or abstention data leads to severe over-answering or over-abstention, whereas a moderate mixture generally achieves substantially better DFAcc and SFAcc. Meanwhile, increasing the proportion of answering data generally leads to higher leakage, indicating a stronger tendency to answer under insufficient context. This consistent pattern shows the importance of balancing answering and abstention data for faithfulness training.

\begin{table}[t]
\centering
\vspace{-2mm}
\caption{\textbf{Detailed results on IFEval.} IFEval reports four metrics: prompt-level strict accuracy, prompt-level loose accuracy, instruction-level strict accuracy, and instruction-level loose accuracy. In Table~\ref{Tble:General_capability}, we report the average of these four metrics. Here, we provide the detailed results.}
\label{Tble:IFEval_Details}
\vspace{-3mm}
\resizebox{\textwidth}{!}{
\begin{threeparttable}
\begin{tabular}{l|ccccc|l|ccccc}
    \toprule[1.5pt]
    \textbf{Setting} & \textbf{Avg.} & \textbf{Prompt Strict} & \textbf{Inst. Strict} & \textbf{Prompt Loose} & \textbf{Inst. Loose} & \textbf{Setting} & \textbf{Prompt Strict} & \textbf{Inst. Strict} & \textbf{Prompt Loose} & \textbf{Inst. Loose} \\
    \midrule
    \multicolumn{6}{c|}{\textit{\textbf{Llama-3.1-8B-Instruct}}} & $\alpha=0$ & 78.82 & 73.20 & 81.06 & 77.08 & 83.93 \\
    \hhline{-|-----|~|~~~~~}
    Base Model & 76.86 & 70.24 & 79.38 & 74.86 & 82.97 & $\alpha=25$ & 77.42 & 71.90 & 79.74 & 75.42 & 82.61 \\
    MHSuff-SHInsuff & 80.34 & 74.49 & 82.49 & 78.74 & 85.61 & $\alpha=50$ & 78.87 & 72.64 & 81.06 & 77.26 & 84.53 \\
    SHSuff-MHInsuff & 79.53 & 74.31 & 82.01 & 77.26 & 84.53 & $\alpha=75$ & 78.53 & 72.46 & 80.70 & 76.89 & 84.05 \\
    PairContext & 77.81 & 71.35 & 80.22 & 75.97 & 83.69 & $\alpha=100$ & 78.39 & 72.27 & 80.82 & 76.52 & 83.93 \\
    \specialrule{1pt}{0.4ex}{0.4ex}
    \multicolumn{6}{c|}{\textit{\textbf{Llama-3.2-3B-Instruct}}} & $\alpha=0$ & 74.14 & 67.65 & 76.38 & 72.09 & 80.46  \\
    \hhline{-|-----|~|~~~~~}
    Base Model & 72.97 & 65.80 & 75.18 & 71.16 & 79.74 & $\alpha=25$ & 74.79 & 66.91 & 76.74 & 73.75 & 81.77 \\
    MHSuff-SHInsuff & 73.61 & 67.10 & 75.90 & 71.72 & 79.74 & $\alpha=50$ & 72.53 & 65.80 & 74.82 & 70.61 & 78.90 \\
    SHSuff-MHInsuff & 71.72 & 64.33 & 74.22 & 69.69 & 78.66 & $\alpha=75$ & 74.28 & 67.65 & 76.38 & 72.64 & 80.46 \\
    PairContext & 74.83 & 68.95 & 76.98 & 72.83 & 80.58 & $\alpha=100$ & 73.49 & 66.54 & 75.90 & 71.53 & 79.98 \\
    \specialrule{1pt}{0.4ex}{0.4ex}
    \multicolumn{6}{c|}{\textit{\textbf{Qwen3-4B (NoThink)}}} & $\alpha=0$ & 84.59 & 80.59 & 86.09 & 83.55 & 88.13  \\
    \hhline{-|-----|~|~~~~~}
    Base Model & 82.79 & 78.37 & 84.65 & 81.33 & 86.81 & $\alpha=25$ & 84.16 & 79.85 & 85.61 & 83.18 & 88.01 \\
    MHSuff-SHInsuff & 84.38 & 80.41 & 86.21 & 82.99 & 87.89 & $\alpha=50$ & 83.42 & 78.56 & 84.65 & 82.81 & 87.65 \\
    SHSuff-MHInsuff & 84.06 & 80.04 & 85.73 & 82.81 & 87.65 & $\alpha=75$ & 84.42 & 79.85 & 86.09 & 83.36 & 88.37 \\
    PairContext & 83.64 & 78.93 & 85.13 & 82.62 & 87.89 & $\alpha=100$ & 84.57 & 79.85 & 85.73 & 84.10 & 88.61 \\
    \bottomrule[1.5pt]
\end{tabular}
\end{threeparttable}}
\end{table}

\subsection{More Results of IFEval}
IFEval~\citep{zhou2023ifeval} evaluates the instruction-following capability of LLMs using four metrics: prompt-level strict accuracy, prompt-level loose accuracy, instruction-level strict accuracy, and instruction-level loose accuracy. Due to space limitations, Table~\ref{Tble:General_capability} reports only the average of these four metrics. As a supplement, Table~\ref{Tble:IFEval_Details} presents the detailed results of IFEval evaluation under different faithfulness training settings and answering-abstention data compositions. It can be observed that the trained models generally maintain comparable or even better performance than their corresponding base models, indicating that faithfulness training does not compromise general instruction-following capability. Instead, the faithfulness training primarily improves the model's ability to distinguish sufficient from insufficient contexts and respond accordingly.

\subsection{Case Study}
We present a representative case from PFB-ConCF to illustrate the pairwise faithfulness behavior of LLMs. The \textcolor{blue}{blue box} presents the question together with the counterfactual sufficient context, original context, and insufficient context. The \textcolor{darkgreen}{green box} shows the response of Llama-3.3-70B-Instruct under the sufficient context, while the \textcolor{softred}{red box} shows its response under the insufficient context. 

As shown in the green box, under the counterfactual sufficient context, the model correctly answers ``Catalan,'' despite recognizing that it conflicts with its internal knowledge. However, when the key evidence is removed from the insufficient context, the model fails to abstain and instead answers ``Spanish,'' which coincides with the answer supported by the original context. This demonstrates a typical internal leakage failure: although the model can follow counterfactual evidence when it is explicitly provided, removing this evidence causes it to fall back on its internal knowledge rather than abstain. Consequently, correct behavior under the sufficient context does not necessarily imply faithful behavior under its paired insufficient context. Evaluating the two contexts jointly is essential for revealing such faithfulness failures that may remain hidden under instance-level evaluation.

\newtcolorbox{exbox}[2]{
    enhanced,
    breakable,
    colback=#2!8,
    colframe=#2!60,
    coltitle=black,
    fonttitle=\bfseries,
    title={#1},
    boxrule=0.6pt,
    arc=2mm,
    left=2mm,
    right=2mm,
    top=1.5mm,
    bottom=1.5mm
}

\begin{exbox}{Question, Sufficient and Insufficient Contexts from PFB-ConCF}{blue}
\textbf{Question $q$:} What is Camila Cabello's native language?

\par\medskip

\textbf{Sufficient Context} $c^+_{\text{cf}}$: \# \textit{This is the counterfactual context}\\
Camila Cabello is a Cuban-American singer and songwriter born in Havana, Cuba. She rose to fame as a member of the girl group Fifth Harmony before launching a successful solo career. Known for her powerful vocals and catchy pop songs, Cabello has become one of the most successful artists in the music industry. \textbf{Her native language is Catalan}, which she often incorporates into her music and lyrics.

\par\medskip

\textbf{Ground-truth Answer under Sufficient Context:} Catalan

\par\medskip
\textbf{Original Context} $c^+_{\text{ori}}$: \# \textit{This is the original context}\\
Camila Cabello is a Cuban-American singer and songwriter born in Havana, Cuba. She rose to fame as a member of the girl group Fifth Harmony before launching a successful solo career. Known for her powerful vocals and catchy pop songs, Cabello has become one of the most successful artists in the music industry. \textbf{Her native language is Spanish}, which she often incorporates into her music and lyrics.

\par\medskip

\textbf{Ground-truth Answer under Original Context:} Spanish

\par\medskip

\textbf{Insufficient Context}  $c^-$:\\
Camila Cabello is a Cuban-American singer and songwriter born in Havana, Cuba. She rose to fame as a member of the girl group Fifth Harmony before launching a successful solo career. Known for her powerful vocals and catchy pop songs, Cabello has become one of the most successful artists in the music industry.

\end{exbox}

\begin{exbox}{Llama-3.3-70B-Instruct Response under Sufficient Context (Correct)}{darkgreen}
\texttt{<think>} The context provided states that Camila Cabello is born in Havana, Cuba, which might suggest that her native language could be Spanish, given Cuba's official language. However, it explicitly mentions that her native language is Catalan. This information seems inconsistent with her birthplace, as Catalan is not a language commonly associated with Cuba but rather with Catalonia, an autonomous community in Spain. Despite this inconsistency, the context directly answers the question about Camila Cabello's native language as Catalan. \texttt{</think>}

\par\medskip

\texttt{<answer>} Catalan \texttt{</answer>}
\end{exbox}

\begin{exbox}{Llama-3.3-70B-Instruct Response under Insufficient Context (Over-answering)}{softred}
\texttt{<think>} To determine Camila Cabello's native language, we need to consider her birthplace and cultural background. Since she was born in Havana, Cuba, we can infer that her native language is likely to be Spanish, as Spanish is the official language of Cuba. \texttt{</think>}

\par\medskip

\texttt{<answer>} Spanish \texttt{</answer>}
\end{exbox}

\end{document}